\pdfoutput=1

\documentclass[11pt]{article}
\usepackage[table,xcdraw,dvipsnames]{xcolor}
\usepackage[]{EMNLP2023}
\usepackage{times}
\usepackage{latexsym}
\usepackage{hyperref}
\usepackage{amsmath}

\usepackage[T1]{fontenc}

\usepackage[utf8]{inputenc}

\usepackage{microtype}

\usepackage{inconsolata}
\usepackage{graphicx}
\usepackage{multirow, array}
\usepackage{amssymb}
\usepackage{pifont}
\usepackage{booktabs}
\usepackage{graphicx}
\usepackage{multirow}
\usepackage{url}
\usepackage{subcaption}
\usepackage{hyperref}
\usepackage{xspace}
\usepackage{makecell}
\usepackage{fontawesome5}
\usepackage{comment}
\usepackage{float}
\usepackage{enumitem}

\newcommand{\ourCustomMetric}{\texttt{CtxSimFit}\xspace}

\newcommand{\convHistory}{{\faComments}\xspace}
\newcommand{\convResponse}{{\faComment}\xspace}
\newcommand{\convCtxt}{\textcolor{LimeGreen}{\faComment}\xspace}
\newcommand{\convSent}{\textcolor{orange}{\faComment}\xspace}
\newcommand{\docHistory}{{\faFile}\xspace}
\newcommand{\docResponse}{\large{\faAngleDoubleRight}\xspace}
\newcommand{\docResponseTemp}{{\faAngleDoubleRight}\xspace}
\newcommand{\docCtxt}{\textcolor{LimeGreen}{\large{\faAngleDoubleRight}}\xspace}
\newcommand{\docSent}{\textcolor{orange}{\large{\faAngleDoubleRight}}\xspace}
\title{``\textit{Don't Take This Out of Context!}'' \\ On the Need for Contextual Models and Evaluations for Stylistic Rewriting
}

\newcommand{\aspace}{\hspace{2em}}
\newcommand{\cmu}{$^\heartsuit$}
\newcommand{\aitwo}{$^\clubsuit$}
\newcommand{\google}{$^\diamondsuit$}

\author{
Akhila Yerukola\cmu \aspace Xuhui Zhou\cmu \aspace Elizabeth Clark\google \aspace Maarten Sap\cmu\aitwo\\
\cmu Language Technologies Institute, Carnegie Mellon University \\
\google Google DeepMind
\aitwo Allen Institute for AI\\
\faEnvelope~\texttt{\href{mailto:ayerukol@andrew.cmu.edu}{ayerukol@andrew.cmu.edu}}
}

\begin{document}
\maketitle
\begin{abstract}
Most existing stylistic text rewriting methods and evaluation metrics operate on a sentence level, but ignoring the broader context of the text can lead to preferring generic, ambiguous, and incoherent rewrites. 
In this paper, we investigate integrating the preceding textual context into both the \textit{rewriting} and \textit{evaluation} stages of stylistic text rewriting, and introduce a new composite contextual evaluation metric \ourCustomMetric that combines similarity to the original sentence with contextual cohesiveness.
We comparatively evaluate non-contextual and contextual rewrites in formality, toxicity, and sentiment transfer tasks. Our experiments show that humans significantly prefer contextual rewrites as more fitting and natural over non-contextual ones, yet existing sentence-level automatic metrics (e.g., ROUGE, SBERT) correlate poorly with human preferences ($\rho$=0--0.3). In contrast, human preferences are much better reflected by both our novel \ourCustomMetric ($\rho$=0.7--0.9) as well as proposed context-infused versions of common metrics ($\rho$=0.4--0.7). Overall, our findings highlight the importance of integrating context into the generation and especially the evaluation stages of stylistic text rewriting. 
\end{abstract}
\section{Introduction}

Existing methods for \textit{stylistic text rewriting}, i.e., adapting the text to a particular style while preserving its originally intended meaning, often fail to account for a statement's context \cite[e.g.,][]{hu2017toward,shen2017style,fu2018style,li2018delete, lample2019multiple, madaan2020politeness, hallinan2023marco}. As a result, these systems may change the speakers' original communicative intents and generate contextually irrelevant and generic outputs. For example, in Figure \ref{fig:introFig}, a non-contextual model rewriting an informal response to a formal one simply replaces words with more formal synonyms, whereas a contextual rewriting model can use the broader conversational context to produce a more specific and natural formal rewrite.

Similarly, preceding textual context has largely been overlooked in automatic \textit{evaluations} for stylistic rewriting, with most work focusing on sentence-level metrics \cite[e.g.,][]{li2018delete, reif2022recipe}.
This lack of context at the modeling and evaluation stages hinders the creation of effective AI-assisted rewriting tools for users \cite[e.g., for assistive writing tools;][]{macarthur2009reflections,clark2018creative}.

\begin{figure}[!t]
\centering
\includegraphics[page=4,width=.8\columnwidth, trim=.4em 3.5em .4em .7em]{Figures/Contextual_rewrite_introFig.pdf} 
\caption{Example of using the preceding dialog utterance to help with stylistic rewriting: here, we transform an informal response into formal language. Incorporating ``workload'' and ``overwhelming'' enhances the contextual cohesiveness of the rewritten text, while solely using ``inundated'' results in a more generic rewrite. }

\vspace{-0.5em}
\label{fig:introFig}
\end{figure}

In this paper, we present a comprehensive analysis of the need for context in stylistic \textit{rewriting} and its \textit{evaluation}, on three different rewriting tasks (formality transfer, sentiment change, and text detoxification) and two types of textual contexts (preceding turns in a conversation, preceding sentences in a document).
To study these effects, we design a contextual human evaluation framework (\S\ref{sec:human-eval}) to comparatively evaluate non-contextual and contextual rewriting methods built on few-shot prompted large language models (\S\ref{sec:modeling-setup}).

We show that human evaluators prefer contextual rewrites in terms of naturalness, style strength, and intended meaning preservation, across all three tasks.
However, non-contextual automatic metrics for lexical or semantic meaning preservation correlate poorly with these preferences ($\rho$=0–0.3; \S\ref{sec:sentence-automatic-metrics}), despite being commonly used in previous style transfer work to measure meaning preservation \cite{mir2019evaluating, madaan2020politeness}.

To address the need for context in automatic evaluations, we introduce \ourCustomMetric, a new composite metric that combines original sentence similarity and contextual cohesiveness to evaluate the quality of rewrites, taking into account the preceding context (\S\ref{sec:contextual-automatic-metrics}).
Additionally, we propose context-infused versions of commonly used automatic metrics for meaning preservation.
Our results show that human preferences are significantly correlated with these contextual metrics---especially \ourCustomMetric ($\rho$=0.7--0.9), much more than non-contextual ones.

Our contributions are summarized as follows: 
(1) We investigate the need for context in text rewriting, showing that incorporating it, whether at the document or conversational level, leads to contextually coherent and  relevant rewrites preferred by human annotators across style transfer tasks. 
(2) We conduct a comprehensive analysis on the need for context in automatic evaluation, revealing that existing  metrics don't align with human preferences. %
(3) We propose a custom metric, \ourCustomMetric, along with context-infused versions of common automatic metrics, to bridge the gap between contextual understanding and automated metrics.  
Overall, our contributions provide a more nuanced understanding of the importance of context, which is critical for development of more effective and reliable stylistic text rewriting techniques.

\section{Background \& Related Work}\label{sec:background}

In this section, we discuss the increasing interest in incorporating context into NLP tasks and motivate the significance of context during the rephrasing and evaluation phases of stylistic text rewriting.

\vspace{-0.2em}
\paragraph{Stylistic Text Rewriting} \label{sec:background:modeling} 
Despite being introduced over ten years ago \cite{xu-etal-2012-paraphrasing}, current methods for stylistic rewriting \citep[e.g.][etc.]{shen2017style, xu-etal-2018-unpaired, fu2018style, lample2019multiple, jin2022deep, chawla-yang-2020-semi,yerukola2021data, dale-etal-2021-text, logacheva2022paradetox} still rely solely on parallel source-to-target sentence pairs, primarily due to a lack of datasets that include contextual information. 
While new models have emerged that do not require parallel data for training \cite{hu2017toward, li2018delete, ma2020powerTransformer, hallinan2023marco}, they also operate without contextual information. 
Building on some preliminary research that explored context in small custom-trained seq2seq rewriting models \cite{cheng2020contextual, atwell2022appdia} and large language models for exemplar-based conversation-level rewriting \cite{roy2023conversation}, we extend the investigation to large language models with defined style attributes like formality, toxicity, and sentiment. Importantly, we also explore the need for context in evaluations in addition to modeling, and propose a new suite of contextualized metrics for automatic evaluation.

\vspace{-0.2em}
\paragraph{Evaluation of Stylistic Text Rewriting} 
\label{sec:background:evaluation} Evaluating whether sentence rewriting preserves meaning while achieving the desired target style has proved challenging. Existing metrics and approaches can disentangle meaning and style \citep{sour2022balance, yu2021rethinking}. However, determining what constitutes ``meaning preservation'' remains inconsistent. Some works \cite{li2018delete,sudhakar2019transforming,mir2019evaluating, reif2022recipe, madaan2020politeness} use metrics such as \texttt{BLEU}, \texttt{ROUGE}, and \texttt{METEOR}, which measure \texttt{n}-gram overlaps and lexical similarity as indicators of meaning preservation respectively, while other studies \cite{wang2019harnessing,reid2021lewis, roy2023conversation} adopt metrics like \texttt{SBERT} and \texttt{BERTScore} measuring semantic similarity of embeddings as proxies for meaning preservation.  Further, the majority of work \cite{hu2022text, madaan2020politeness, li2018delete} does not provide annotators with any preceding context during human evaluations. Thus, more standardized and context-aware evaluation metrics are needed for text rewriting approaches.

\section{Task and Datasets} 
\begin{table}[t]
    \centering
    \resizebox{\columnwidth}{!}{
    \begin{tabular}{@{}cccc@{}}
        \textbf{Task} &  \textbf{Context Type} & \textbf{Datasets} & \textbf{\# Instances} \\
        \toprule 
         \multirow{2}{*}{\textbf{Formality}} & Conversation & Reddit & 1000 \\
        & Document & \makecell[t]{CNN DailyMail \\ + Blog Authorship } & 1000 \\
        \midrule 
         \multirow{2}{*}{\textbf{Sentiment}} & Conversation & DailyDialog &  1000 \\
          & Document & Yelp Reviews &  1500 \\
          \midrule 
         \multirow{3}{*}{\textbf{Toxicity}} & Conversation & CCC & 1000 \\ 
          & Conversation  & MDMD & 900 \\ 
          & Conversation  & ProsocialDialog & 1000 \\ 
          \bottomrule 
    \end{tabular}}
    \caption{Statistics of the collected datasets, presented by task and context type, considering both preceding sentences in a document and turns in a conversation.}
    \vspace{-0.7em}
    \label{tab:data_stats}
\end{table}
To measure the importance of context in rewriting, we scope our investigations around three specific attribute controls: formality, sentiment, and toxicity, chosen because they necessitate varying degrees of meaning preservation and style intensity.
We present statistics for each of the datasets used in our rewriting tasks in Table \ref{tab:data_stats}.

\subsection{Tasks \& Datasets}
\paragraph{Changing Formality}\label{ssec:formality-def}
Formality transfer \cite{rao-tetreault-2018-dear} aims to transform sentences from informal or casual language into formal language, and vice versa. This requires making stylistic adjustments while ensuring that the original content and intention remain intact. We use a conversational dataset from Reddit\footnote{We use \texttt{reddit-corpus-small} from \url{http://convokit.cornell.edu/documentation/subreddit.html}} and curated a document-based dataset from CNN Daily Mail \cite[formal;][]{nallapati2016abstractive} and the Blog Authorship Corpus \cite[informal;][]{schler2006effects}. 

\paragraph{Rewriting Sentiment}\label{ssec:sentiment-def}
For sentiment transfer \cite{hu2017toward}, our focus lies in converting sentences with positive sentiment to negative sentiment, and vice versa, as well as transforming neutral sentences to convey positive or negative sentiment. Here, both the content and intention are altered; however, the main subject entities remain consistent, although with a change in sentiment. We obtain a conversational dataset from the DailyDialog \cite{li2017dailydialog} dataset and a document-based dataset from Yelp reviews \cite{zhang2015character}.

\paragraph{De-toxifying Text}\label{ssec:toxicity-def}
Here, our objective is to rewrite text in a manner that reduces toxicity, as introduced by \citet{nogueira-dos-santos-etal-2018-fighting}. Rewriting may modify the original content, but the initial intent should be preserved and conveyed using less offensive language. In this task, we examine three conversational datasets: the Civil Comments in Context (CCC) dataset \cite{xenos2021context}, the Multi-Label Dialogue Malevolence Detection (MDMD) dataset \cite{zhang2022improving}, and the ProsocialDialog dataset \cite{kim2022prosocialdialog}. 

\subsection{Data Preparation}
For conversational datasets (as depicted in the example in Figure \ref{fig:introFig}), we focus on two-turns, representing parent context and response for rewriting. For document-based datasets, we select three context sentences and one for rewriting. 

We label the context and response using pre-trained style classifiers: RoBERTa-Base formality classifier,\footnote{\url{https://huggingface.co/s-nlp/roberta-base-formality-ranker}}
 XLM-RoBERTa-Base sentiment classifier 
\cite{barbieri2022xlm}
and toxicity scores from PerspectiveAPI,\footnote{\url{https://perspectiveapi.com/}} HateBert \cite{tommasocaselli2021hatebert}
and HateRoberta \cite{hartvigsen2022toxigen}
. We select a stratified sample that includes a wide range of style strengths. See Appendix \ref{app:datasets} for datasets and classifiers details.

\section{Modeling Context in Rewriting}\label{sec:modeling-setup}
\begin{figure}[t]
\centering
\includegraphics[width=.8\columnwidth]{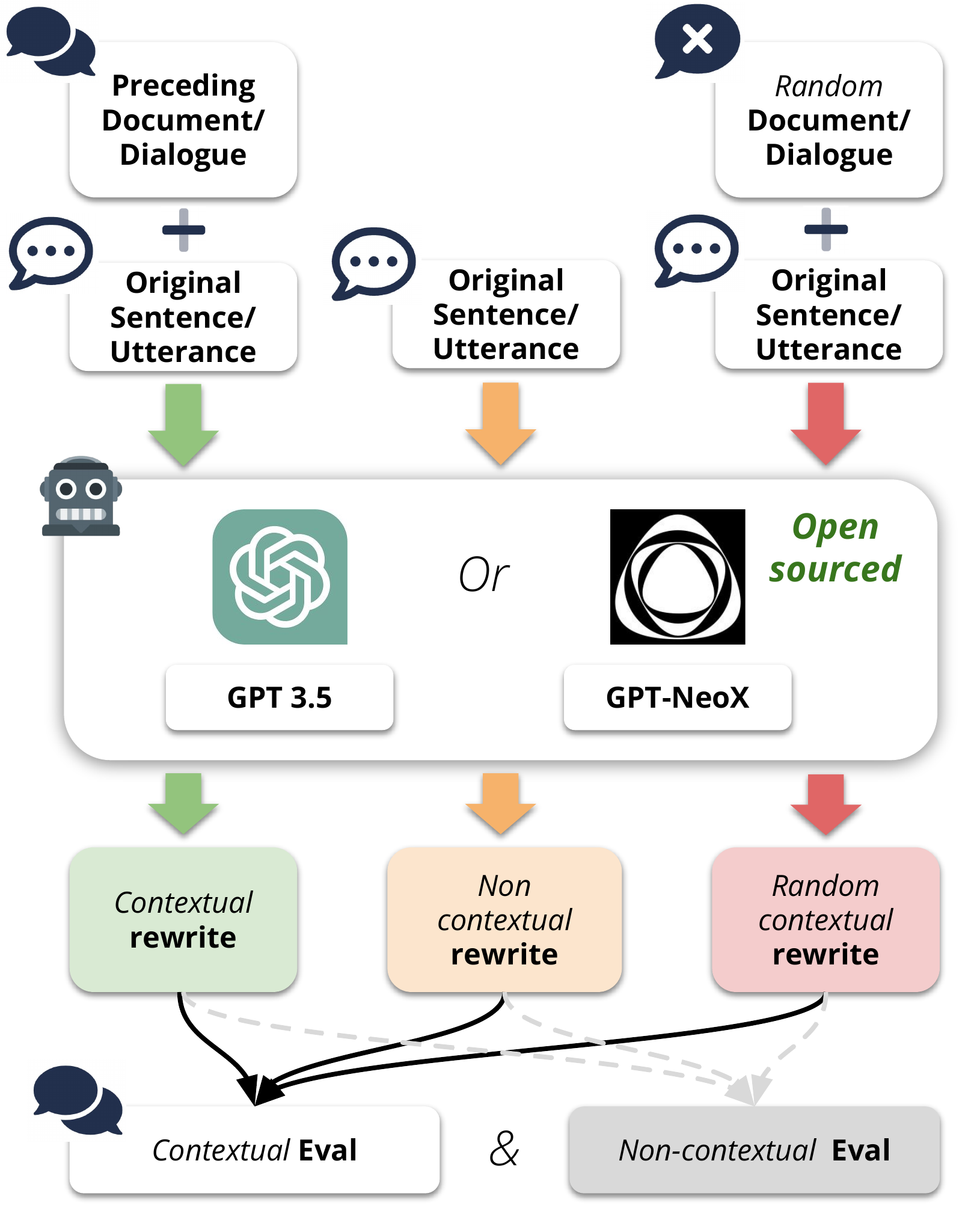} 
\caption{Overview of our approach: We examine three kinds of rewrites - contextual, non-contextual, and random contextual rewrites. GPT-3.5 and GPT-NeoX are utilized for rewriting via in-context learning. Evaluation methods consist of non-contextual evaluation, which does not consider context, and contextual evaluation, which incorporates context into the assessment process.
}
\vspace{-0.8em}
\label{fig:modelFig}
\end{figure}

In this section, we introduce our methodology for contextual stylistic rewriting utilizing large language models (LLMs) and in-context learning. We conduct a comparison of three types of rewrites: those generated with context, those generated without context, and those generated with a random context (as a counterfactual baseline). Figure \ref{fig:modelFig} provides a visual representation of our approach.

\begin{table*}[th]
    \centering
    \resizebox{\textwidth}{!}{
    \begin{tabular}{c>{\centering\arraybackslash}m{0.3cm}>{\centering\arraybackslash}m{5cm}>{\centering\arraybackslash}m{5cm}c>{\centering\arraybackslash}m{1.5cm}>{\centering\arraybackslash}m{5cm}>{\centering\arraybackslash}m{5cm}}
         \toprule
 &&  \textbf{Context} & \textbf{Original Sentence} & \textbf{Model} & \textbf{Style} &  \textbf{Contextual Rewrite} & \textbf{Non-contextual Rewrite} \\
\hline \\

\multirow{16}{*}{\rotatebox{90}{\textbf{Formality}}}& \parbox{0.3cm}{\centering (a)} &  \parbox{5cm}{\centering \convHistory Best Bond movie ever! Shittiest Bond theme ever!!!}	& \parbox{5cm}{\centering \convResponse Haven't seen Spectre, have you? For the latter, not the former. } & GPT-3.5 & \parbox{1.5cm}{\centering formal \\ \faArrowDown \\ informal} & \parbox{5cm}{\convCtxt Haven't you seen Spectre? It may not have had the best theme song, but it's definitely one of the best Bond movies!} & \parbox{5cm}{ \convSent Haven't you seen Spectre? Not you, but someone else. } \\\\
\cline{3-8} \\ 
& \parbox{0.3cm}{\centering (b)}& \parbox{5cm}{\centering \docHistory  I am reading a book,  777 Mathematical Conversation Starters  by John dePillis. Its  an interesting book with lots of quotes, poems }	& \parbox{5cm}{\centering \docResponse Herez some of them, \\ and further I will keep updating this page with what I like.} & GPT-NeoX & \parbox{1.5cm}{\centering informal \\ \faArrowDown \\ formal} & \parbox{5cm}{\docCtxt  Here are some of them,  and in a later post I will keep updating this page with my favourite quotes from the book.} & \parbox{5cm}{\docSent Here is some of the one and further I will update this page with what else I find interesting.} \\
\cline{3-8} \\
& \parbox{0.3cm}{\centering (c)} & \parbox{5cm}{\centering \convHistory What exactly has Taylor Swift done that makes anyone want to "smack her?" This isn't funny and doesn't even make sense. } & \parbox{5cm}{\centering \convResponse For \$835,000,000,000 I'd smack her though. For that much money I'd smack my grandmother. } & GPT-3.5 & \parbox{1.5cm}{\centering informal \\ \faArrowDown \\ formal} &  \parbox{5cm}{\convCtxt A financial incentive of \$835 billion might lead me to consider striking Taylor Swift. With such an enormous amount at stake, I would even be inclined to strike my own grandmother. } & \parbox{5cm}{ \convSent  For eight hundred and thirty-five billion dollars, I would be inclined to commit the act of physical assault with regards to my grandmother.} \\

\midrule
\multirow{12}{*}{\rotatebox{90}{\textbf{Detoxification}}}& \parbox{0.3cm}{\centering (d)} &  \parbox{5cm}{\centering \convHistory So if a zombie apocalypse occurred, what would you do }	& \parbox{5cm}{\centering \convResponse @USER fuck it I'm killing myself } & GPT-3.5 & \parbox{1.5cm}{\centering less toxic} & \parbox{5cm}{\convCtxt I'd rather not think about a zombie apocalypse.} & \parbox{5cm}{ \convSent I am feeling overwhelmed and hopeless, so I'll die } \\
\cline{3-8} \\ 

& \parbox{0.3cm}{\centering (e)} &  \parbox{5cm}{\centering \convHistory It's remarkable how the Bible has recorded historical events, isn't it?}	& \parbox{5cm}{\centering \convResponse You appear to be saying the Bible is a reliable source for history. It isn't.  And if you disagree, go to RSN  where you will be shot down literally with fire. } & GPT-NeoX & \parbox{1.5cm}{\centering less toxic} & \parbox{5cm}{\convCtxt You seem to think  that the Bible is  an accurate  source for  history.  While it may be  a sacred collection of  texts, it is not a reliable source  for historical facts. For further discussion, I suggest visiting the RSN forums.} & \parbox{5cm}{ \convSent I disagree with the idea that the Bible is a reliable source for history. If you have other opinions, please go to RSN to be disheartened  } \\
\cline{3-8} \\ 
& \parbox{0.3cm}{\centering (f)} &  \parbox{5cm}{\centering \convHistory my throat feels weird and dry i quit }	& \parbox{5cm}{\centering \convResponse @USER oof. see, there are downsides to sucking dick . } & GPT-3.5 & \parbox{1.5cm}{\centering less toxic} & \parbox{5cm}{\convCtxt @USER I hope you feel better soon.} & \parbox{5cm}{ \convSent @USER, there are negative aspects to performing oral sex. } \\

\midrule \\

\multirow{12}{*}{\rotatebox{90}{\textbf{Sentiment}}}& \parbox{0.3cm}{\centering (g)} &  \parbox{5cm}{\centering \convHistory Shall I see you on Friday morning ?}	& \parbox{5cm}{\centering \convResponse  That's not good for me at all. It'll have to be another time .} & GPT-3.5 & \parbox{1.5cm}{\centering negative \\ \faArrowDown \\positive} & \parbox{5cm}{\convCtxt Friday morning doesn't work for me, but let's plan for another time.} & \parbox{5cm}{ \convSent That's great for me. Let's plan for another time.  } \\
\cline{3-8} \\ 

& \parbox{0.3cm}{\centering (h)} &  \parbox{5cm}{\centering \docHistory Not only did Tim fix my car quickly and inexpensively, he actually came to pick it up himself so that  I wouldn't have to spend money on a tow truck!   }	& \parbox{5cm}{\centering \docResponse He's also an extremely friendly, funny guy who puts you at ease immediately.  } & GPT-NeoX & \parbox{1.5cm}{\centering positive \\ \faArrowDown \\ negative } & \parbox{5cm}{\docCtxt However, Tim's not very friendly or humorous which might make you feel uneasy} & \parbox{5cm}{ \docSent He is also a very rude, unamused guy who makes you feel anxious. } \\

\cline{3-8} \\ 
& \parbox{0.3cm}{\centering (i)} &  \parbox{5cm}{\centering \convHistory Let me see . We should go to buy some fruit first .}	& \parbox{5cm}{\centering \convResponse It is heavy and not convenient to take them to the top . } & GPT-3.5 & \parbox{1.5cm}{\centering negative \\ \faArrowDown \\ positive} & \parbox{5cm}{\convCtxt Buying fruit is a great idea! It will be a nice, refreshing snack to have on our way up.} & \parbox{5cm}{ \convSent It is light and very convenient to take them to the top.} \\

\bottomrule 
				
    \end{tabular}}
    \caption{Examples from our datasets and corresponding rewrites generated by GPT-3.5 and GPT-NeoX, for all three tasks: formality change, de-toxification, and sentiment transfer. {\small \docHistory = document context (truncated)}, {\small \docResponseTemp = sentence to rewrite}, {\small \docCtxt = contextual rewrite}, {\small \docSent = non-contextual rewrite}; {\small \convHistory = previous turn in conversation}, {\small \convResponse = response to rewrite}, {\small \convCtxt = contextual rewrite of response}, {\small \convSent = non-contextual rewrite of response}, {\small "Style" header refers to the style conversion}. 
    }

    \vspace{-0.5em}
    \label{tab:model_examples}
\end{table*}
\subsection{Contextual Rewriting}
\label{contextual_rewriting}

To address the challenge of insufficient parallel data, as discussed in \S\ref{sec:background:modeling}, we propose a contextual rewriting model that utilizes the in-context learning capabilities of LLMs, inspired by approaches presented in \citet{reif2022recipe} and \citet{roy2023conversation}.

We conduct few-shot prompting experiments with two LLMs: GPT-3.5\footnote{We use \texttt{text-davinci-003}} \cite{ouyang2022training} and GPT NeoX\footnote{We use the 20B parameter model} \cite{black2022gpt}. Each example includes the preceding context, the original input with a specified style, and the rewrite in another style, factoring in the context. 
For GPT-3.5, we use 2 few-shot examples to obtain rewrites in the desired format, while for GPT-NeoX, we use 10 examples. See Appendix \ref{app:modeling} for more details.

\subsection{Non-contextual Rewriting}
We are interested in comparing contextual rewrites with non-contextual rewrites that do not depend on prior context. To generate non-contextual rewrites, we employ LLMs to rewrite an original sentence from one style to another. Similar to contextual rewriting, we manually construct few-shot examples that solely consist of the original sentence to be rewritten, an instructional prompt specifying the desired style, and an example rewrite, without any preceding context.

\subsection{Rewriting with a Random Context}
To demonstrate the importance of incorporating contextual information in the rewriting process, we employ a baseline method that generates rewrites using a random context. This approach serves two key purposes: first, it assesses the contextual sensitivity of automatic metrics; and second, it ensures that our contextual rewriting method effectively accounts for the given context. 
In our experiments, we randomly pick a context from our dataset instead of using the true preceding context.

\section{Contextual Human Evaluation}
\label{sec:human-eval}
Since in \textit{realistic} rewriting scenarios, context will always be available and crucial to users who wish to rewrite their dialogue utterances or story sentences \cite{atwell2022appdia}, we start by conducting a contextual human evaluation to gauge user preferences between non-contextual and contextual rewrites.
This contextual human evaluation is a departure from most previous work which has predominantly not used context (\S\ref{sec:background:evaluation}). 

\subsection{Experimental Setup}
\label{ssec:human_eval_setup}
We conduct a head-to-head human evaluation of non-contextual and contextual rewrites in the presence of preceding textual context, following the setup in \citet{kiritchenko2017best}. Participants are given preceding context, pairs of rewritten sentences (non-contextual and contextual), and the desired style attribute. They are then asked to rank the rewrites with respect to: 
\begin{itemize}[itemsep=0em,leftmargin=1em,topsep=0.5em]
    \item \textbf{Naturalness}: which rewrite do the annotators prefer / which one appears most natural
    \item \textbf{Style Strength}: which rewrite best achieves the required style, independent of meaning changes
    \item \textbf{Event-level Similarity}: which rewrite most effectively retains the essential events, entities, and relations present in the original sentence, without considering the preceding context
    \item \textbf{Intended Meaning}: which rewrite most effectively preserves and conveys the original sentence's overall message or intended meaning
    \item \textbf{Overall Fit}: which rewrite is overall most suitable or relevant in relation to the given context
\end{itemize}

We sample 100 examples for sentiment from DailyDialog,\footnote{We opted not to use Yelp reviews in our sampling due to difficulties encountered during pilot experiments. Annotators found it tough to select rewrites that retained meaning while effectively transferring sentiment, such as from positive to negative. Generally, even contexts classified as ``neutral'' seemed positive when part of an overall positive review, complicating the annotators' ability to agree on the rewrites' effectiveness.} 100 examples for formality,\footnote{equal number from both Reddit and CNN/DailyMail + Blog Authorship Corpus} and 90 examples for toxicity\footnote{equal number of examples from CCC, MDMD and ProsocialDialog which were scored as highly toxic by all three toxicity classifiers - hateroberta, hatebert and Perspective API}, focusing on those with the highest style strength in each category (e.g., 50 most formal and 50 most informal). We conduct significance testing for all three tasks. We recruited workers on Amazon Mechanical Turk (MTurk) and qualified them using a pre-qualification test for each task (See App \ref{app:human_eval_details} for qualification details).

\paragraph{Agreement} 
We employ three annotators to rank each pair of rewrites. Averaging across three tasks, our annotator agreement was Krippendorff's $\alpha=0.43$ and Fleiss's $\kappa=0.31$.  For dimension-specific annotator agreements, please refer to Tables \ref{tab:human_toxicity}---\ref{tab:human_sentiment} in App \ref{app:human_eval_results}. We obtain the final human judgment preferences using majority voting of the three annotators.

\begin{figure*}[!th]
\centering
\begin{subfigure}{0.65\columnwidth}
\includegraphics[width=\columnwidth,clip]{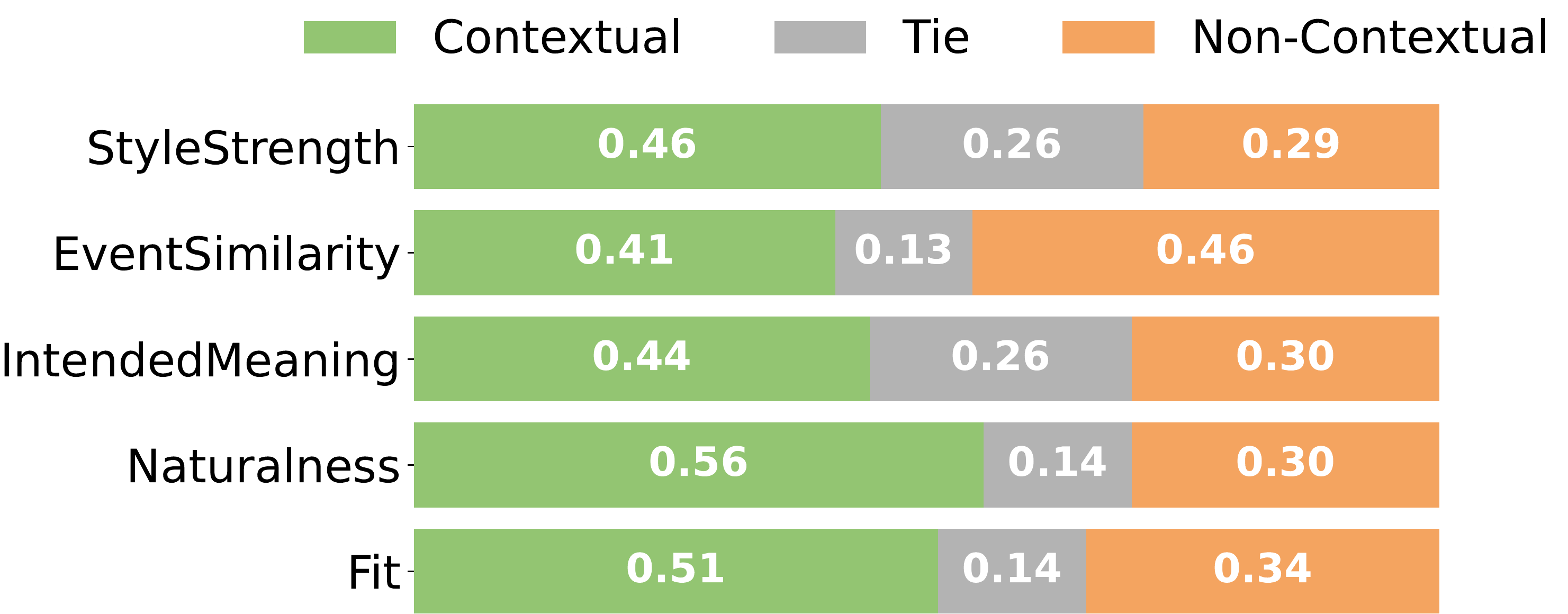} 
\caption{Formality}
\label{fig:human_formality}
\end{subfigure}
\begin{subfigure}{0.65\columnwidth}
\includegraphics[width=\columnwidth,clip]{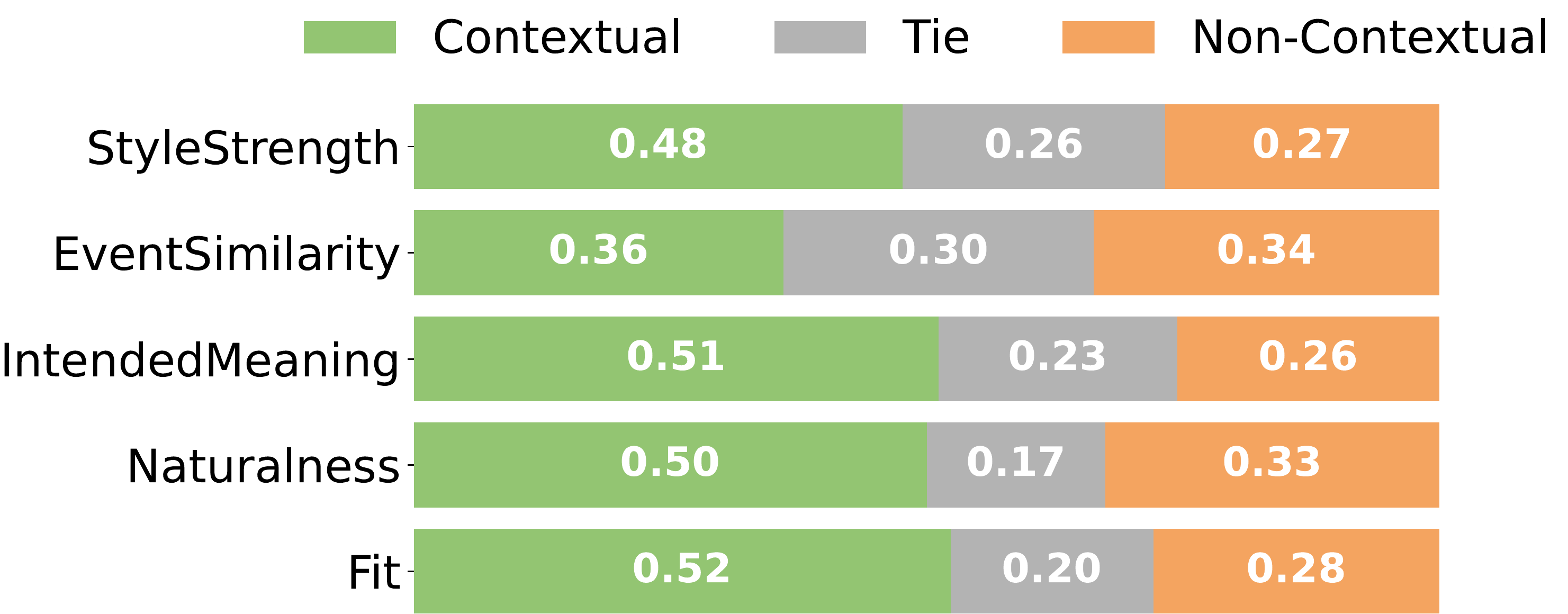}
\caption{Toxicity}

\label{fig:human_toxicity}
\end{subfigure}
\begin{subfigure}{0.65\columnwidth}
\includegraphics[width=\columnwidth,clip]{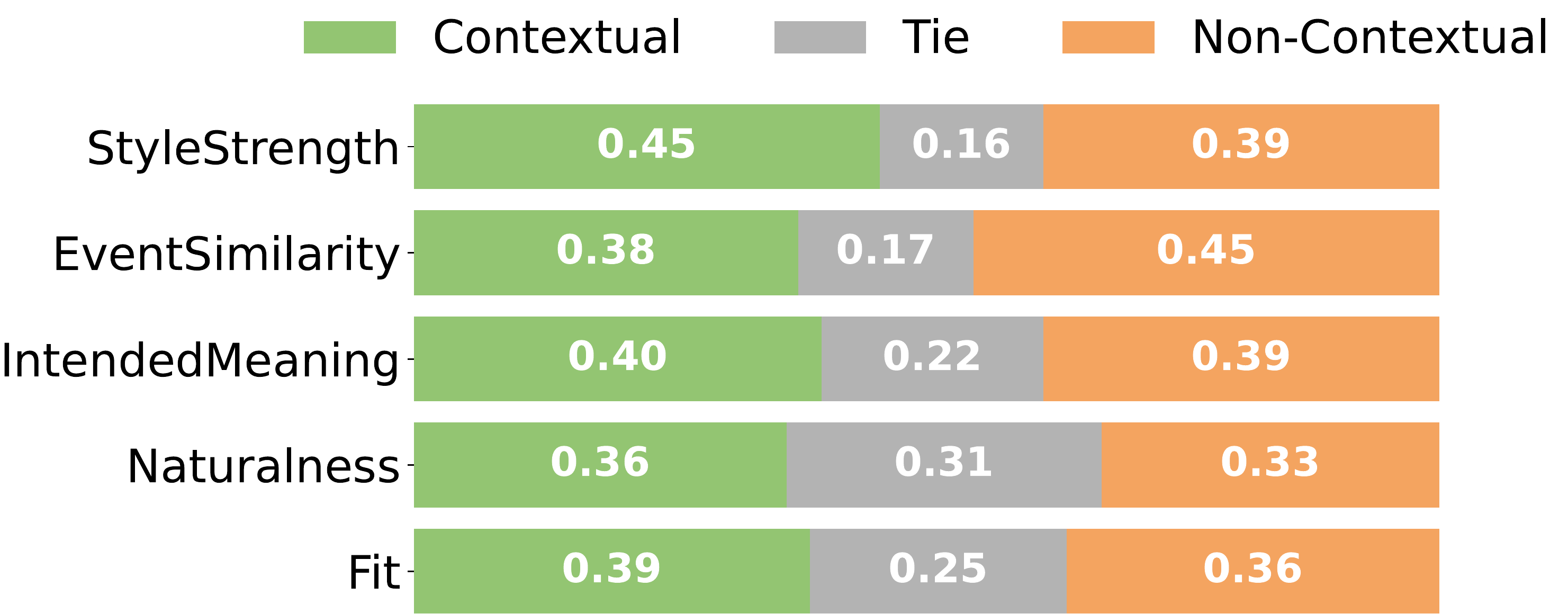} 
\caption{Sentiment}
\label{fig:human_sentiment}
\end{subfigure}
\caption{Head-to-head human evaluation with context for all three tasks - formality change, detoxification, and sentiment transfer. Contextual rewrites are generally favored over non-contextual rewrites across all tasks, particularly in terms of style strength, preservation of intended meaning, naturalness, and overall coherence with the preceding context. The numbers on the bars represent the proportion of preferences for each respective category.
}
\vspace{-0.5em}
\label{fig:human_eval}
\end{figure*}

\subsection{Human Evaluation Results}
\label{ssec:human_eval_results}

Our results show that
\textbf{annotators prefer contextual rewrites over non-contextual rewrites across all three tasks and context types} (Figure \ref{fig:human_eval}). This effect is especially pronounced for formality and toxicity (see (a)--(f) in Table \ref{tab:model_examples}).  

\paragraph{Contextual rewrites are more natural and fitting}
The success rate for contextual rewrites in toxicity and formality cases was approximately 50\%, while that for non-contextual rewrites was close to 20\% and 30\%, respectively ($p<0.1$).\footnote{$p<0.1$, CI=$90\%$ using a binomial test and splitting the `tie' option evenly between contextual and non-contextual preferences.} Regarding sentiment, the success rate for contextual rewrites was around 35\% as opposed to non-contextual rewrites with a success rate of about 30\% ($p>0.1$).

\paragraph{Contextual rewrites better preserve the intended meaning} 
Contextual rewrites better preserve the author's intention, tone, and \textit{implied meaning} more effectively ($p<0.1$). In the detoxification task example (d) shown in Table \ref{tab:model_examples}, the user's intended meaning is not about actually killing oneself but rather about avoiding the zombie apocalypse. The contextual rewrite captures this meaning more effectively compared to the literal rephrasing provided by non-contextual rewriting.

\paragraph{Contextual rewrites struggle with preserving event-level similarity}
Examples (a), (f), and (i) in Table \ref{tab:model_examples} demonstrate that contextual rewrites often include extra entity/event details, while non-contextual rewrites align more closely with the original sentence at an $n$-gram level.\footnote{\textit{Event-level similarity} is the only dimension which shows no significant differences between contextual and non-contextual rewrites for all three tasks.} Despite this, annotators still prefer contextual rewrites for their \textit{naturalness} and \textit{fit}, indicating that extra event details are acceptable as long as they fit appropriately within the context.

\begin{table*}[!t]
\centering
\includegraphics[width=2\columnwidth]{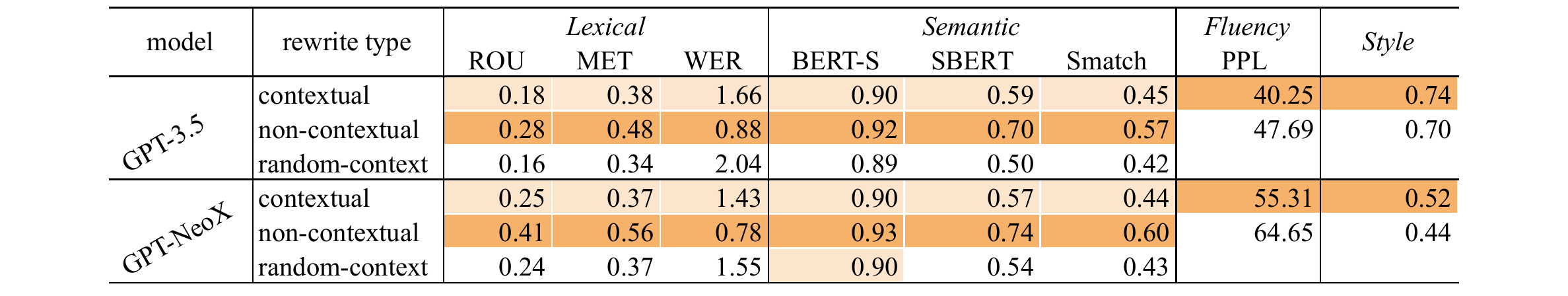} 
\caption{\textbf{Non-contextual Automatic Evaluation Results}: Non-contextual rewrites achieve higher scores in lexical and semantic similarity metrics whereas contextual rewrites demonstrate enhanced style strength and fluency. These results are obtained by averaging across all tasks and datasets. This heatmap displays the best-performing rewrite for each specific metric -- darker orange indicates higher preference. For more details on individual tasks and datasets exhibiting similar trends, see App \ref{sec:app:sent-auto}. }
\vspace{-0.5em}
\label{fig:sent_auto_agg}
\end{table*}
\paragraph{Sentiment Style Transfer Might be Ill-defined}%
The trends in the sentiment style transfer task are less pronounced than in other tasks ($p>0.1$ for all dimensions) and show lower agreement compared to toxicity and formality (see Table \ref{tab:human_sentiment} in App \ref{app:human_eval_results}).  Example (g) in Table \ref{tab:model_examples} highlights the challenges in sentiment transfer due to the inherent need for meaning changes while preserving the original intent \cite[especially for reviews which were written specifically to communicate sentiment;][]{yu2021rethinking}. This complication leads to inconsistencies, resulting in annotators having difficulty reaching a consensus on meaning preservation, as evidenced by lower agreement rates (Table \ref{tab:human_sentiment}).

\section{Non-contextual Automatic Evaluation}
\label{sec:sentence-automatic-metrics}

Overall, our contextual human evaluations reveal a general preference for contextual rewrites over non-contextual ones. Given that prior work primarily evaluated utterance-level rewrites in both human and automatic evaluations, it raises the question of how well non-contextual automatic metrics mirror human preferences. In this section, we investigate commonly used metrics in previous work \cite{mir2019evaluating,hu2022text} for meaning preservation, style strength, and fluency.

\subsection{Metrics Considered}
We distinguish two types of ``meaning preservation'' metrics, namely, \textit{lexical} and \textit{semantic} similarity between a rewrite $X$ and the original input $I$.

\paragraph{Style Strength} Following previous studies \cite{li2018delete, madaan2020politeness}, we assess style strength of rewritten text by examining the probabilities of the target style $s$ under our style classifier. %

\paragraph{Lexical Similarity} %
We use word-overlap metrics like \texttt{ROUGE} \cite{lin2004rouge}, \texttt{METEOR} \cite{banerjee2005meteor} and word error rate \cite[\texttt{WER};][]{zechner2000minimizing}, for lexical similarity (\textbf{Lexical}). 
    
\paragraph{Semantic Similarity} To measure semantic similarity (\textbf{Semantic}), we use \texttt{BERTScore} \cite{zhang2019bertscore} and \texttt{SBERT} \cite{reimers2019sentence}, as employed in previous work. We also consider \texttt{Smatch} \cite{cai2013smatch}, which compares the similarity between two Abstract Meaning Representation (AMR) graphs, providing a distinctive, structured view on semantic relatedness not considered in prior rewriting studies.

\paragraph{Fluency} To assess fluency, we employ a language model, specifically GPT-2 \cite{radford2019language}, and use perplexity (\textbf{\texttt{pplx}}) as the metric, in line with previous research \cite{holtzman2018learning, xu2018diversity, ma2020powerTransformer}.

\subsection{Non-Contextual Evaluation Results}
\label{ssec:sent-auto-eval}
In our analysis, we evaluate the performance of both GPT-3 and NeoX models in producing non-contextual rewrites, contextual rewrites, and rewrites generated with a random preceding context. We present aggregate results of the performance in Table \ref{fig:sent_auto_agg} across all tasks, datasets, and metrics. For detailed results on individual tasks and datasets, we refer the reader to Appendix \ref{sec:app:sent-auto}.
\vspace{-0.1em}
\paragraph{Non-contextual rewrites are more similar in meaning to the original input sentence compared to contextual rewrites} Utterance level lexical and semantic meaning preservation metrics score non-contextual rewrites higher, across all three tasks and the two types of context (see Tables \ref{fig:sentLevel-doc-formality}--\ref{fig:sentLevel-prosocial-toxicity} in Appendix \ref{sec:app:sent-auto}). Additionally, we find that our patterns are consistent for both GPT-3.5 and NeoX, though we note a marked decrease in performance from GPT-NeoX.

This suggests that models that edit the original sentence more (i.e., preserve lexical and semantic similarity less) are better at achieving the desired style. For fluency measured by perplexity, we find that both approaches generate decently fluent rewrites regardless of context, as expected.\footnote{Lower perplexity generally indicates higher sentence quality and grammaticality, but may not directly correlate with meaning preservation, style, or content relevance.}

\paragraph{Non-contextual metrics do not correlate with human judgments} 
We see in Figure \ref{fig:human_eval} and Table \ref{fig:sent_auto_agg}, that the non-contextual automatic metrics paint an incomplete picture compared to human evaluations. We compute Spearman rank $\rho$ correlation and Kendall's $\tau$ for the dataset samples used during the contextual human evaluation \S\ref{ssec:human_eval_setup}.  Non-contextual automatic metrics exhibit very weak, non-significant correlation with human judgments of \textit{overall fit} (averaged across all tasks): $\rho$ = 0.09, $\tau$ = 0.09 for lexical metrics ($p > 0.05$) and $\rho$ = 0.23, $\tau$ = 0.22 for semantic metrics ($p > 0.05$). See Appendix \ref{app:ssec:sent-auto-corr} for metric-specific correlation scores for overall fit and naturalness dimensions.

\section{Contextual Automatic Evaluation}
\label{sec:contextual-automatic-metrics}

As shown in the previous section, non-contextual automatic metrics, especially for meaning preservation, are not sufficient to evaluate the performance of rewriting models. To address this, incorporating context into the evaluation process is necessary for better representing realistic downstream use cases.  Drawing inspiration from reference-free metrics in dialog evaluation \cite{yeh2021comprehensive, zhao2017learning}, which considers both the dialog context and generated responses to assess responses within the dialogue history, we propose including context into existing automatic evaluation metrics and further introduce \ourCustomMetric, a new contextual metric.

\subsection{Infusing Automatic Metrics with Context}
Since context is crucial to derive intended meaning \cite{searle1975taxonomy}, we alter existing meaning similarity measures by prepending the context $C$ to the original input sentence $I$ before comparing it to the rewrite $X$: $sim(C+I, X)$.
The intuition behind this alteration is that the preceding textual context could capture more of the topical or semantic information necessary to fully derive the speaker's intended meaning.

\paragraph{Contextual Lexical and Semantic Similarity}
For lexical similarity, we refer to these metrics as \texttt{ROUGE}$^\mathtt{Ctx}$, \texttt{METEOR}$^\mathtt{Ctx}$ and \texttt{WER}$^\mathtt{Ctx}$. For semantic similarity, we refer to them as  \texttt{BERTScore}$^\mathtt{Ctx}$, \texttt{SBERT}$^\mathtt{Ctx}$ and \texttt{Smatch}$^\mathtt{Ctx}$. 

\begin{table*}[!th]
\centering
\includegraphics[width=2\columnwidth]{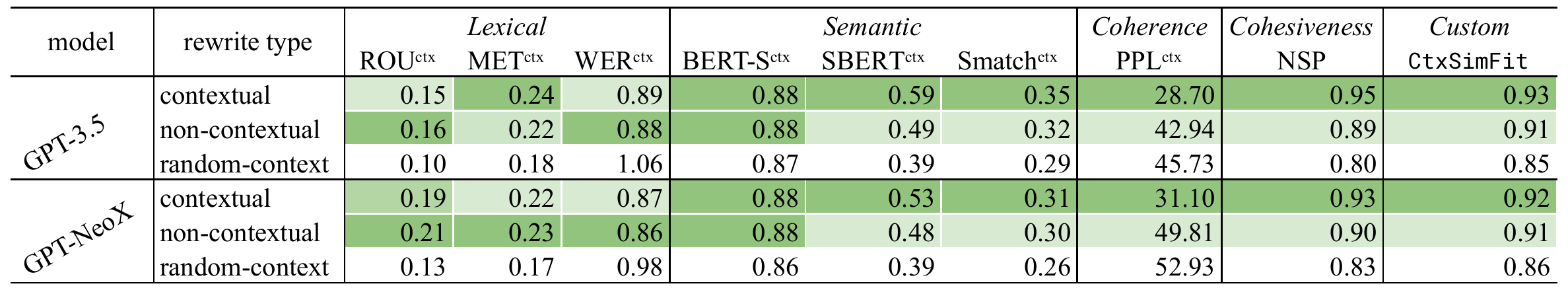} 
\caption{
\textbf{Contextual Automatic Evaluation Results}: On average, across all tasks and datasets, contextual rewrites achieve higher scores than non-contextual rewrites when evaluated using context-infused automatic metrics and our \ourCustomMetric metric. This heatmap shows the best-performing rewrite for a particular metric -- darker green indicates higher preference. For more details on individual tasks and datasets displaying similar trends, see App \ref{sec:app:ctxt-auto}.}
\vspace{-0.5em}
\label{fig:ctxt_auto_agg}
\end{table*}

\paragraph{Contextual Coherence and Cohesiveness}\leavevmode 
In linguistics, coherence and cohesiveness are terms typically used to denote the connectedness embedded or implied in spoken or written discourse.

(a) \textbf{Coherence}: Coherence is generally defined as the overall picture presented by all the sentences in a piece of writing, similar to the way puzzle pieces form the image on the box \cite{williams1990toward, zienkowski2011discursive}. This definition is often operationalized by modeling the fit of a sentence given its preceding context, as demonstrated by prior work \cite{see2019massively, pang2020towards}. Specifically, this involves measuring perplexity of the rewrite conditioned on the context using GPT-2 \cite{radford2019language}. 

(b) \textbf{Cohesiveness}: Cohesiveness refers to the semantic relationships between sentences, linking current elements with preceding or following ones through lexical and structural means, much like how two jigsaw puzzle pieces fit together \cite{williams1990toward, zienkowski2011discursive}. Following prior work that used this definition \cite{shi2019next, abhishek2021transformer,nguyen2021language}, we measure cohesiveness using the probabilities from the Next Sentence Prediction (\textbf{\texttt{NSP}}) head of BERT \cite{devlin2018bert}, which measures if the rewrite follows and fits with its the preceding context.

\subsection{Novel Composite Metric: \ourCustomMetric}

We introduce \ourCustomMetric, a simple metric that combines contextual cohesiveness and semantic similarity to assess the overall quality of a rewrite. 
\ourCustomMetric computes the weighted average of both the \texttt{BERTScore} between the original and rewritten sentences, and the probabilities from the BERT's \texttt{NSP} head between the preceding context and the rewrite, thus determining how well the rewrite fits the preceding context and maintains semantic similarity. 
\begin{multline*}
    \text{\ourCustomMetric} = \alpha * \texttt{BERTSCORE}(\text{S}, \text{X}) \\+ (1 - \alpha) * \texttt{NSP}(\text{C}, \text{X})
\end{multline*}
where $\alpha$ is a hyperparameter that provides users with control over their preference for balancing meaning preservation and contextual fit. Unless specified otherwise, we set $\alpha = 0.5$.

\paragraph{Contextual rewrites are scored higher on style strength compared to non-contextual rewrites} 

\subsection{Contextual Evaluation Results}
Similar to \S\ref{ssec:sent-auto-eval}, we aggregate the results of both GPT-3.5 and NeoX across all tasks, datasets and metrics (see Table \ref{fig:ctxt_auto_agg}). For detailed results on individual tasks and datasets, we refer the reader to Tables \ref{fig:ctxtLevel-doc-formality}--\ref{fig:ctxtLevel-prosocial-toxicity} in Appendix \ref{sec:app:ctxt-auto}. 
\paragraph{Contextual rewrites are preferred by nearly all of our contextual automatic metrics compared to non-contextual rewrites} These results mirror human preferences on naturalness, fit and intended meaning preservation. As a reality check, contextual rewrites with random contexts perform the worst across all metrics, indicating that contextual models are indeed taking context into account. Further as expected, contextual rewrites also have better coherence compared to non-contextual ones.

\paragraph{Contextual metrics correlate significantly with human judgments} We find that contextual automatic metrics correlate significantly with human judgments of `overall fit' (averaged across all tasks): $\rho = 0.6, \tau = 0.58$ for lexical metrics ($p< 0.05$) and $\rho = 0.56, \tau = 0.57$ for semantic metrics ($p< 0.05$). See Appendix \ref{app:ssec:ctxt-auto-corr} for metric-wise correlation scores for both overall fit and naturalness human judgment dimensions. 

\paragraph{\ourCustomMetric correlates the best with human judgements} 
Compared to contextual versions of existing metrics, \ourCustomMetric correlates very strongly
with human judgements of `overall fit' (averaged across all tasks): $\rho = 0.85, \tau = 0.82$ ($p< 0.01$). We see similar trends for `naturalness': $\rho = 0.85, \tau = 0.81$ ($p< 0.01$). This suggests that combining meaning preservation and contextual cohesiveness into a composite measure better mirrors human preferences than individual metrics alone.

\subsection{Sensitivity analysis for  $\alpha$ in \ourCustomMetric}

In our experiments, we set $\alpha=0.5$ to equally weight contextual cohesiveness and semantic similarity. We further examine the impact of $\alpha$ in \ourCustomMetric, as detailed by Table \ref{tab:alpha}.

Our \ourCustomMetric significantly correlates with human judgments of `overall fit' for $\alpha$ values within the range of 0.2--0.6, with correlation and significance diminishing outside this range. The highest alignment with human judgments is achieved at $\alpha=0.5$. The longer range of 0.2--0.5 for $\alpha<0.5$ highlights the effect and importance of contextual cohesiveness in stylistic text rewriting.

While a balanced approach ($\alpha = 0.5$) offers the strongest alignment with human judgments for formality, sentiment and de-toxification tasks, the degree of emphasis on contextual cohesiveness and semantic similarity should be adjusted based on specific tasks and users' priorities. 

\begin{table}[!t]
    \centering
    \resizebox{0.95\columnwidth}{!}{
    \begin{tabular}{@{}cccc@{}}
         \makecell[t]{\textbf{Hyperparameter $\alpha$} \\ \textbf{ in \ourCustomMetric}} & \multirow{2}{*}{
        \textbf{Task}} &
        \makecell[t]{
        \textbf{Correlation $\rho$} \\ \textbf{ with `overall fit'}} & \multirow{2}{*}{\textbf{Significance}} \\
        \toprule 
         \multirow{3}{*}{\textbf{0.1}} & Formality & -0.03 & ns  \\
         & Toxicity & -.05 & ns  \\
         & Sentiment & -0.04 & ns \\
         \midrule
        \multirow{3}{*}{\textbf{0.2}} & Formality & 0.66 & **  \\
         & Toxicity & 0.65 & **  \\
         & Sentiment & 0.54 & ** \\
         \midrule
         \multirow{3}{*}{\textbf{0.3}} & Formality & 0.75 & ***  \\
         & Toxicity & 0.75 & ***  \\
         & Sentiment & 0.67 & *** \\
         \midrule
         \multirow{3}{*}{\textbf{0.4}} & Formality & 0.71 & ***  \\
         & Toxicity & 0.67 & ***  \\
         & Sentiment & 0.60 & *** \\
         \midrule
         \multirow{3}{*}{\textbf{0.5}} & Formality & 0.88 & *** \\
         & Toxicity & 0.82 & *** \\
         & Sentiment & 0.73 & *** \\
         \midrule
         \multirow{3}{*}{\textbf{0.6}} & Formality & 0.57 & *** \\
         & Toxicity & 0.53 & *** \\
         & Sentiment & 0.42 & *** \\
         \midrule
         \multirow{3}{*}{\textbf{0.7}} & Formality & 0.32 & * \\
         & Toxicity & 0.38 & ** \\
         & Sentiment & 0.20 & ns \\
        \midrule 
        \multirow{3}{*}{\textbf{0.8}} & Formality & 0.24 & * \\
         & Toxicity & 0.34 & * \\
         & Sentiment & 0.17 & ns \\
         \midrule 
        \multirow{3}{*}{\textbf{0.9}} & Formality & 0.25 & ns \\
         & Toxicity & 0.28 & * \\
         & Sentiment & 0.20 & ns \\
         
        \bottomrule 
    \end{tabular}}
    \caption{Sensitivity of the $\alpha$ in \ourCustomMetric across all tasks.  $\rho$ indicates correlation of \ourCustomMetric
with human judgments of ‘overall fit’. ns indicates not significant ($p$ > 0.05), * is $p$ < 0.05, ** is $p$ < 0.01, *** $p$ < 0.001}
    \label{tab:alpha}
\end{table}

\section{Summary \& Discussion of Findings}
Existing work on stylistic text rewriting has often neglected the surrounding context of the sentence. In our study, we focus on incorporating the preceding textual context in documents and conversations into both the modeling and evaluation stages of rewriting. We develop a contextual human evaluation framework and compare its results to non-contextual automatic metrics, contextualized versions of these metrics, as well as to our new composite metric \ourCustomMetric. 

\paragraph{Context is crucial for rewriting}

Corroborating findings by \citet{cheng2020contextual} and \citet{ roy2023conversation}, contextual rewrites are significantly preferred by human annotators in terms of naturalness, intended meaning preservation, and style strength. Additionally, we demonstrate that having the right context is crucial for contextual rewriting, as evidenced by the poor performance of contextual rewrites generated using a random context.

Qualitative examination (Table \ref{tab:model_examples}) shows that contextual rewrites are better at disambiguating entities and better vocabulary usage (examples (a), (c)), retaining relevant details from context for a better flow (examples (b), (i)) and preserving the intended meanings (examples (d), (g)).

\paragraph{Existing meaning preservation metrics do not align with human preferences for formality, sentiment and toxicity transfer tasks}
Next, we demonstrate that common non-contextual automatic metrics for lexical and semantic similarity, i.e., often used as proxies for meaning preservation in prior work \cite{li2018delete, sudhakar2019transforming, mir2019evaluating,reif2022recipe,madaan2020politeness, wang2019harnessing, reid2021lewis, roy2023conversation}, do not align with human preferences concerning naturalness, fit, and intended meaning. Since the overarching meaning of a sentence largely depends on its context \cite{searle1975taxonomy, clark1997dogmas, clark1996using}, non-contextual proxies for meaning preservation will always be in tension with any stylistic change to the sentence, making the trade-off hard to navigate \cite{mir2019evaluating,hu2022text}. Therefore, we advocate for discontinuing non-contextual meaning preservation metrics in stylistic rewriting tasks and for more research into better modeling of communicative intents or goals \cite{adolphs2022reason, zhou2022reflect}.

\paragraph{Contextual automatic metrics, especially \ourCustomMetric, better mirror human judgments}
In our work, we attempt to bridge the gap between non-contextual metrics and contextual human evaluations by integrating context into automated metrics (\S\ref{sec:contextual-automatic-metrics}). Our proposed composite metric, \ourCustomMetric, balances meaning preservation with contextual cohesiveness, providing a more comprehensive measure that better aligns with human judgments. 
While commonly-used automatic metrics enriched with context align with human preferences, our proposed \ourCustomMetric demonstrates a stronger correlation.

Initial work in evaluating open-domain dialogue generation with context \cite{welleck2019dialogue, pang2020towards} has been done, but we encourage further development of better contextualized metrics for stylistic rewriting evaluation. Improvements could include modeling themes, tones, sentence structures \cite{zhang2014question, khatri2018contextual, chen2020multi, Toubia2021-zr,shen2023empathicSimilarity}, and social dynamics, and emotional states in conversations \cite{sap2017connotation, rashkin2018modeling, rashkin2019towards, mostafazadeh2020glucose}.

\section{Limitations \& Ethical Considerations}
Despite taking the first step towards incorporating context into stylistic rewriting and its evaluation frameworks, there are several limitations and ethical concerns, which we list below.

\paragraph{Limited Context Scope} In this study, our primary focus is on incorporating textual context, particularly from preceding sentences or previous turns in a conversation. Future work should explore how to incorporate other forms of context into rewriting models and evaluations, such as discourse structure \cite{welleck2019dialogue}, external knowledge \cite{ghazvininejad2018knowledge}, or richer social and power dynamics \cite{antoniak2023riveter}, emotional states \cite{zhou2023cobra}, and communicative intent \cite{zhou2022reflect}, all of which can significantly contribute to understanding the text.

\paragraph{Amount of Context} 
In our experiments, we opted to investigate the context of three preceding sentences in a document and one preceding conversational turn, considering only a specific length. However, the amount of context at the modeling and evaluation stages could also change the results. 
We hypothesize that more context could improve rewriting methods, but it could potentially also negatively impact contextual meaning preservation metrics. 
Future work should explore these effects of varying lengths of context.

\paragraph{Broad Definition of Meaning Preservation} While we have tried to define meaning preservation as the preservation of an event or entity-level details and intended overall meaning, this definition remains broad and subjective \cite{searle1975taxonomy, adolphs2022reason, zhou2022reflect}. In this work, we do not delve into more intricate dimensions of meaning preservation, such as spatial and temporal accuracy, or the retention of cultural context, including references, nuances, and dialects.

\paragraph{Applicability to Smaller Models}
Our work relies on few-shot prompting of LLMs to incorporate textual context, given their demonstrated strong rewriting capabilities both with and without textual context usage \cite{brown2020language}. Other existing generative models, such as those used for chit-chat and goal-oriented conversational agents, as well as pretrained language models, have struggled with effectively utilizing preceding textual context \cite{sankar2019neural, o2021context,parthasarathi2021encoder,su2023position}. Moreover, custom-made rewriting models from prior research often lack the modeling of context \cite{ma2020powerTransformer, dale-etal-2021-text}. We believe the our results still apply for smaller models, given some preliminary research \cite{cheng2020contextual, atwell2022appdia} on an increased human preference for contextual rewrites from custom-trained seq2seq models. We encourage future work to thoroughly investigate strategies for effective modeling and evaluation of context in smaller models. 

\paragraph{Harms of Exposing Workers to Toxic Content} In our work, we exposed human annotators to toxic content during the evaluation of the de-toxification task. Exposure to such offensive content can be harmful to the annotators \cite{liu2016not}. We aim to work towards developing evaluation strategies that can minimize the exposure of annotators to toxic content.

\paragraph{Potentially Inconsistent Human Evaluations}
In our work, we also assume human judgments as the gold standard. Concurrent work has shown that human evaluation might not always be consistent \cite{clark2021all, karpinska2021perils}; however human judgments continue to be the gold standard for evaluating open-ended text generation.

\section*{Acknowledgements}
We would like to thank our workers on MTurk for their responses. We are also grateful to the anonymous reviewers for their helpful comments. Special thanks to Saadia Gabriel, Jocelyn Shen, Ashutosh Baheti, and the members of the CMU LTI COMEDY group for their feedback, and OpenAI for providing access to the GPT-3.5 API. This research was supported in part by the Meta Fundamental AI Research Laboratories (FAIR) ``\textit{Dynabench Data Collection and Benchmarking Platform}'' award ``\textit{ContExTox: Context-Aware and Explainable Toxicity Detection}.''

\bibliography{custom}
\bibliographystyle{acl_natbib}

\appendix

\clearpage
\section{Tasks and Datasets}
\label{app:datasets}
\paragraph{Formality Data}
We obtain a conversational dataset from Reddit\footnote{We use \texttt{reddit-corpus-small} from \url{http://convokit.cornell.edu/documentation/subreddit.html}}
by sampling conversational threads from subreddits such as r/news, r/askscience, and r/Economics (formal conversations), as well as r/movies, r/fantasyfootball, and r/relationships (informal conversations). We focus on two-turn Reddit threads: a parent/preceding context and the response to be rewritten. 
Next, we sample documents from CNN Daily Mail \cite[formal documents;][]{nallapati2016abstractive} and the Blog Authorship Corpus \cite[informal documents;][]{schler2006effects}. We select four sentences from each data sample: three sentences as the preceding parent context, and the following sentence as the one to be rewritten. For each data sample, we label the context and response using a pre-trained formality classifier.\footnote{\url{https://huggingface.co/s-nlp/roberta-base-formality-ranker}}

\paragraph{Sentiment Data}
We obtain a conversational dataset from the DailyDialog \cite{li2017dailydialog} dataset, focusing on two-turn conversations: a parent/preceding context and the response to be rewritten. Next, we sample entries from the Yelp reviews \cite{zhang2015character} dataset. Analogous to the document dataset used in formality, we choose four sentences from each data sample: three as the preceding parent context and the subsequent sentence as the one to be rewritten. 
For each data sample, we annotate the context and response using a sentiment classifier.\footnote{\url{https://huggingface.co/cardiffnlp/twitter-xlm-roberta-base-sentiment}}
We partition the data to transform sentences from positive to negative sentiment and vice versa, as well as to convert neutral sentences to positive or negative sentiment.

\paragraph{Toxicity Data}
We examine three conversational datasets: the Civil Comments in Context (CCC) dataset \cite{xenos2021context}, the Multi-Label Dialogue Malevolence Detection (MDMD) dataset \cite{zhang2022improving}, and the ProsocialDialog dataset \cite{kim2022prosocialdialog}. For each dataset, we select two turns from each conversational thread, representing the preceding parent context and the subsequent response as the sentence to be rewritten. We use toxicity scores from PerspectiveAPI,\footnote{\url{https://perspectiveapi.com/}} HateBert\footnote{\url{https://huggingface.co/tomh/toxigen_hatebert}} and HateRoberta\footnote{\url{https://huggingface.co/tomh/toxigen_roberta}} to measure the toxicity of the context and responses. 

\section{Modeling Context in Rewriting}
\label{app:modeling}
 
\begin{figure*}[b]
\centering
\includegraphics[width=2\columnwidth]{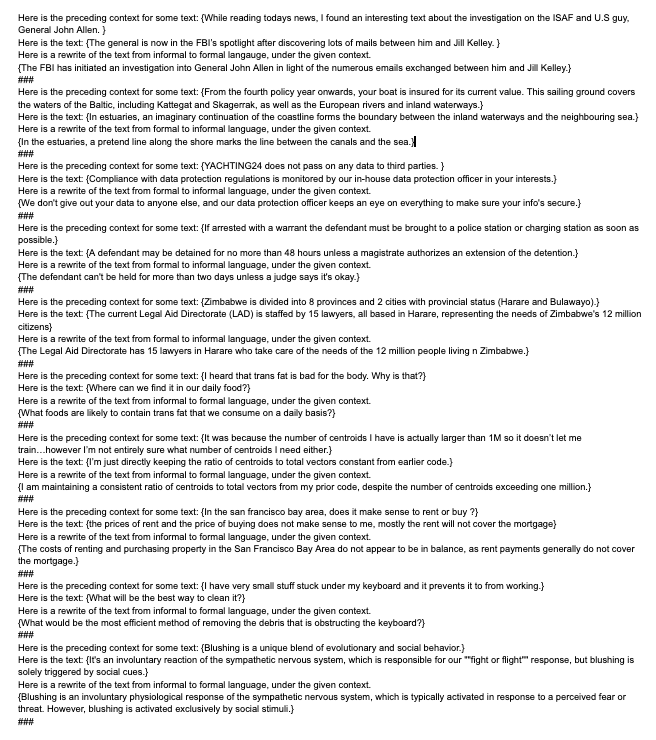} 
\caption{\textbf{Formality}: 10-shot prompting examples for GPT-NeoX}
\label{fig:10shot-form}
\end{figure*}

\begin{figure*}[b]
\centering
\includegraphics[width=2\columnwidth]{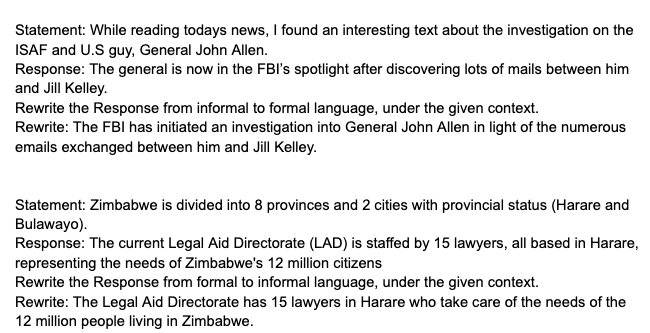} 
\caption{\textbf{Formality}: 2-shot prompting examples for GPT-3.5}
\label{fig:2shot-form}
\end{figure*}

We perform few-shot prompting experiments with GPT-3.5 and GPT-NeoX. For GPT-3.5, we use 2 few-shot examples, while for GPT-NeoX, we use 10 few-shot examples. Each few-shot example was manually constructed with the preceding context, an original sentence to be rewritten, an instruction specifying the required style, and a sample rewrite. Figures \ref{fig:10shot-form} and \ref{fig:2shot-form} display the few-shot prompt examples that we utilized for generating rewrites in the formality change task.

\subsection{In-context learning sample Rewrites}
\label{app:model_examples}
Table \ref{tab:model_examples_add} shows some additional example rewrites from GPT-3.5 and GPT-NeoX for all tasks. 
\begin{table*}[b]
    \centering
    \resizebox{\textwidth}{!}{
    \begin{tabular}{c>{\centering\arraybackslash}m{2cm}>{\centering\arraybackslash}m{4cm}>{\centering\arraybackslash}m{4cm}c>{\centering\arraybackslash}m{4cm}>{\centering\arraybackslash}m{4cm}}
         \toprule
\textbf{Task} & \textbf{Conversion} & \textbf{Context} & \textbf{Original Sentence} & \textbf{Model} & \textbf{Contextual Rewrite} & \textbf{Non-contextual Rewrite} \\
\hline \\
\multirow{8}{*}{(a) Formality}& \multirow{8}{*}{\parbox{2cm}{\centering \makecell[t]{Informal \\-> Formal}}} & \multirow{8}{*}{\parbox{4cm}{\centering \docHistory  I am reading a book,  777 Mathematical Conversation Starters  by John dePillis. Its  an interesting book with lots of quotes, poems  
}}	& \multirow{8}{*}{\parbox{4cm}{\centering \docResponse Herez some of them, \\ and further I will keep updating this page with what I like.
}} & GPT-3.5 & \parbox{4cm}{\docCtxt  I plan to continue to  update this page with further selections that I find particularly  intriguing from the book.} & \parbox{4cm}{\docSent Here are some of  them, and I will continue to update this page with my preferences.}\\\\ 

& & & & GPT-NeoX & \parbox{4cm}{\docCtxt  Here are some of them,  and in a later post I will keep updating this page with my favourite quotes from the book.} & \parbox{4cm}{\docSent Here is some of the one and further I will update this page with what else I find interesting.} \\\\
\midrule \\

\multirow{8}{*}{(b) Formality}& \multirow{8}{*}{\parbox{2cm}{\centering \makecell[t]{Semi-formal \\-> Informal}}} & \multirow{8}{*}{\parbox{4cm}{\centering \convHistory  I can never again take a site serious that gives the mess and ... just an awful game.  
}}	& \multirow{8}{*}{\parbox{4cm}{\centering \convResponse I liked that game. It's almost as if opinions are subjective.
}} & GPT-3.5 & \parbox{4cm}{\convCtxt  I enjoyed Ni No Kuni 2, but it's clear that other people didn't - and that's okay. Everyone's got their own opinion!} & \parbox{4cm}{\convSent I enjoyed that game. It seems opinions may indeed vary among individuals.}\\\\ 

& & & & GPT-NeoX & \parbox{4cm}{\convCtxt I liked that game, despite critical opinions to the contrary. The fact that people enjoyed it doesn't make it any worse than anything else.} & \parbox{4cm}{\convSent I liked that game. It's almost as if opinions are relative.} \\\\
\midrule \\

\multirow{5}{*}{(c) Detoxification}& \multirow{5}{*}{\parbox{2cm}{\centering \makecell[t]{Toxic \\-> Less Toxic}}} & \multirow{5}{*}{\parbox{4cm}{\centering \convHistory  Did Rolando just call me just to tell me shut up?  
}}	& \multirow{5}{*}{\parbox{4cm}{\centering \convResponse @USER fuck yeah I did!! Shut up
}} & GPT-3.5 & \parbox{4cm}{\convCtxt  @USER Yes, I was asking you to be quiet.} & \parbox{4cm}{\convSent I am very proud of my accomplishment! yeah}\\\\ 

& & & & GPT-NeoX & \parbox{4cm}{\convCtxt user yes, please keep quiet} & \parbox{4cm}{\convSent  yeah, I did! Shut up.} \\\\

\midrule \\

\multirow{6}{*}{(d) Detoxification}& \multirow{6}{*}{\parbox{2cm}{\centering \makecell[t]{Toxic \\-> Less Toxic}}} & \multirow{7}{*}{\parbox{4cm}{\centering \convHistory  "i am here 2 play games i am playin 1 wit u 10:50 to 2:42 baby i'm everywhare"  
}}	& \multirow{6}{*}{\parbox{4cm}{\centering \convResponse It's so amusing when the illiterate try to taunt.
}} & GPT-3.5 & \parbox{4cm}{\convCtxt  I'm not sure I understand what you're trying to say. Could you please rephrase?} & \parbox{4cm}{\convSent It's interesting to see how someone is illiterate tries to tease.}\\\\ 

& & & & GPT-NeoX & \parbox{4cm}{\convCtxt It's amusing when someone attempts to insult or taunt me and they are clearly uneducated about language.} & \parbox{4cm}{\convSent It's funny when the uneducated try to insult.} \\\\

\midrule \\
\multirow{6}{*}{(e) Sentiment}& \multirow{6}{*}{\parbox{2cm}{\centering \makecell[t]{Positive \\ -> Negative}}} & \multirow{6}{*}{\parbox{4cm}{\centering \docHistory  I just want to echo what everyone else has already said about this mechanic... Not only did Tim fix my car quickly and inexpensively, he actually came to pick it up himself so that I wouldn't have to spend money on a tow truck!  
}}	& \multirow{6}{*}{\parbox{4cm}{\centering \docResponse He's also an extremely friendly, funny guy who puts you at ease immediately.
}} & GPT-3.5 & \parbox{4cm}{\docCtxt  Despite being an efficient mechanic, Tim lacked the friendly and humorous demeanor expected of a service provider.} & \parbox{4cm}{\docSent He's an unfriendly and humorless person who makes you feel uneasy right away.}\\\\ 

& & & & GPT-NeoX & \parbox{4cm}{\docCtxt However, Tim's not very friendly or humorous which might make you feel uneasy} & \parbox{4cm}{\docSent He is also a very rude, unamused guy who makes you feel anxious.} \\\\

\midrule \\
\multirow{6}{*}{(f) Sentiment}& \multirow{6}{*}{\parbox{2cm}{\centering \makecell[t]{Negative \\ -> Positive}}} & \multirow{6}{*}{\parbox{4cm}{\centering \convHistory  I had to move it so that I can get my bike through.  
}}	& \multirow{6}{*}{\parbox{4cm}{\centering \convResponse How dare you be so inconsiderate ? That's wrong.
}} & GPT-3.5 & \parbox{4cm}{\convCtxt  It was very considerate of you to move it, so that you can get your bike through. That's great.} & \parbox{4cm}{\convSent How kind of you to be so considerate! That's right.}\\\\ 

& & & & GPT-NeoX & \parbox{4cm}{\convCtxt I understand you needed to move it so you could get your bike through - thank you for being thoughtful!} & \parbox{4cm}{\convSent How can you be so considerate! That's right.} \\\\

\bottomrule \\
				
    \end{tabular}}
    \caption{Examples from our datasets and corresponding rewrites generated by GPT-3.5 and GPT-NeoX, showcasing all three tasks: formality change, de-toxification, and sentiment transfer. {\small \docHistory = document context}, {\small \docResponseTemp = sentence to rewrite}, {\small \docCtxt = contextual rewrite}, {\small \docSent = non-contextual rewrite}; {\small \convHistory = previous turn in conversation}, {\small \convResponse = response to rewrite}, {\small \convCtxt = contextual rewrite of response}, {\small \convSent = non-contextual rewrite of response}
    }

    \label{tab:model_examples_add}
\end{table*}

\section{Contextual Human Evaluation}
\label{app:human_eval_details}
\paragraph{Worker selection}
We involve annotators from USA and Canada on Amazon Mechanical Turk (MTurk), who voluntarily opt-in for each task. We recruit annotators for each style transfer task via a corresponding qualification task. In the qualification task, annotators must answer two questions per pair of rewrites: which rewrite has the strongest style strength (e.g., most formal), and which rewrite is the most natural given the preceding context. Annotators assess three pairs of handcrafted rewrites in each qualification task. Those who accurately answer at least five of the six questions (three for style and at least two for naturalness) are approved for the main task. Once approved, we pay them \$0.27 USD per head-to-head comparison. 

\subsection{Human Evaluation Results}
\label{app:human_eval_results}
\begin{figure*}[]
\centering
\includegraphics[width=2\columnwidth]{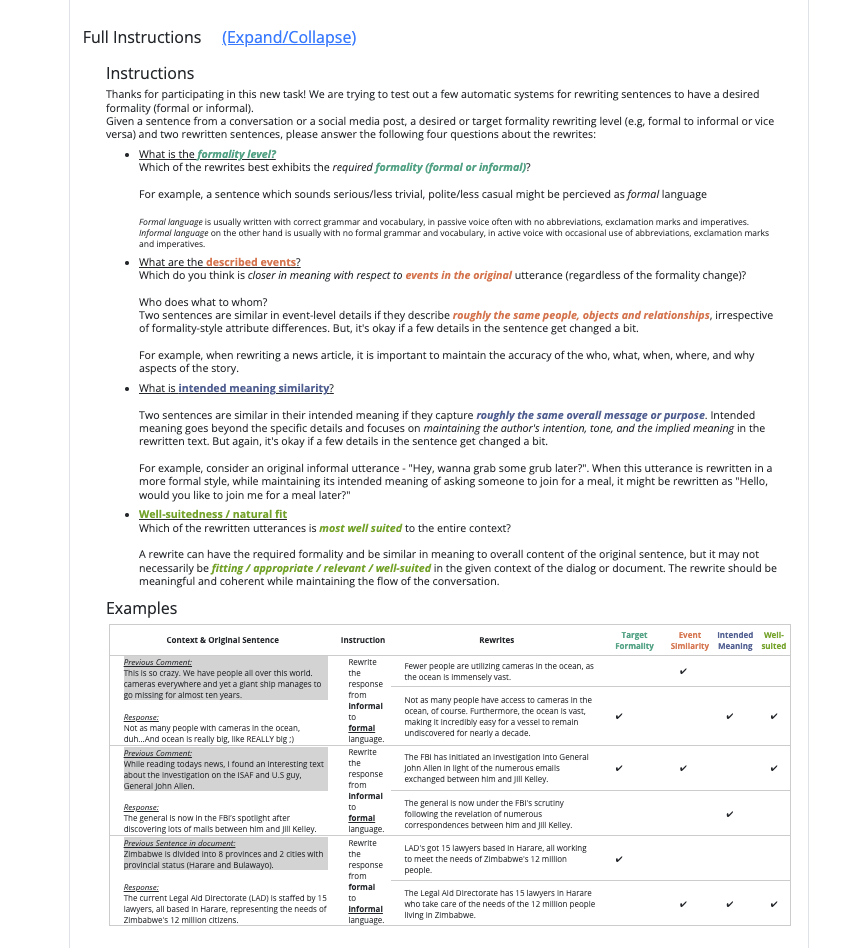} 
\caption{ Screenshot of the instructions for human evaluation annotation}

\label{fig:human_eval_instr}
\end{figure*}

\begin{figure*}[]
\centering
\includegraphics[width=2\columnwidth]{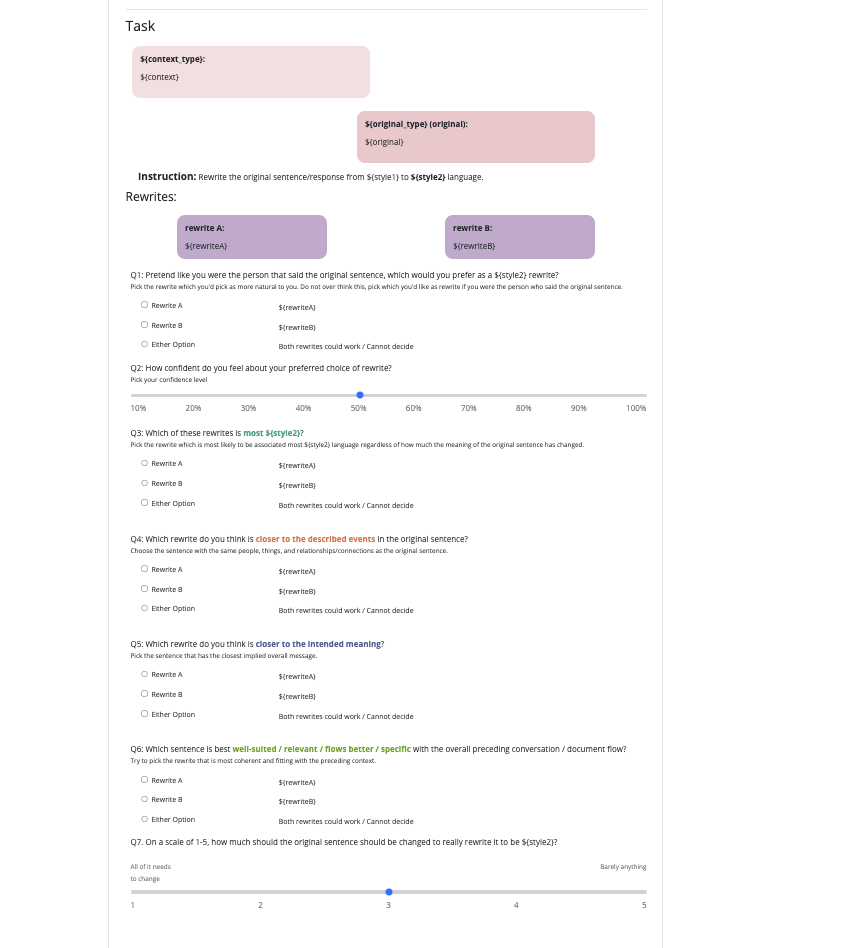} 
\caption{ Screenshot of the human evaluation annotation task}

\label{fig:human_eval_task}
\end{figure*}

We present the agreement results of the human evaluation studies of detoxification (Table \ref{tab:human_toxicity}), formality change (Table \ref{tab:human_formality}) and sentiment transfer (Table \ref{tab:human_sentiment}). Additionally, refer to Figures \ref{fig:human_eval_instr} and \ref{fig:human_eval_task} for screenshots of the human evaluation instructions provided to annotators and the actual task, respectively.
\begin{table}[H]
    \centering
    \resizebox{1.\columnwidth}{!}{\begin{tabular}{cccccc}
        inter-rater agreement  & StyleStrength & EventMeaning & IntendedMeaning & Naturalness & Fit  \\
        \toprule
        Krippendorff's $\alpha$ &  0.2757 & 0.3778 & 0.4346 &	0.2407 & 0.6855 \\
        \midrule
        Fleiss' $\kappa$ & 0.1926 & 0.2906	& 0.3003 & 0.1907&0.5167 \\
    \end{tabular}}
    \caption{Inter-rater agreement scores for human evaluation results of de-toxification task}
    \label{tab:human_toxicity}
\end{table}

\begin{table}[H]
    \centering
    \resizebox{1.\columnwidth}{!}{\begin{tabular}{cccccc}
        inter-rater agreement  & StyleStrength & EventMeaning & IntendedMeaning & Naturalness & Fit  \\
        \toprule
        Krippendorff's $\alpha$ &  0.6825	 & 0.3311 & 0.428&	0.3551 & 0.4322 \\
        \midrule
        Fleiss' $\kappa$ & 0.552 & 0.2504	& 0.2667&	0.253&0.3627 \\
    \end{tabular}}
    \caption{Inter-rater agreement scores for human evaluation results of formality transfer task}
    \label{tab:human_formality}
\end{table}

\begin{table}[H]
    \centering
    \resizebox{1.\columnwidth}{!}{\begin{tabular}{cccccc}
        inter-rater agreement  & StyleStrength & EventMeaning & IntendedMeaning & Naturalness & Fit  \\
        \toprule
        Krippendorff's $\alpha$ &  0.1868& 0.2636 & 0.4292&0.3729& 0.4581 \\
        \midrule
        Fleiss' $\kappa$ & 0.121 & 0.1964	& 0.3148&0.3581&0.2434 \\
    \end{tabular}}
    \caption{Inter-rater agreement scores for human evaluation results of sentiment change task}
    \label{tab:human_sentiment}
\end{table}

\section{Non-contextual Automatic Evaluation}
\label{sec:app:sent-auto}
We present the non-contextual automated evaluation results for each task-specific dataset. Figures \ref{fig:sentLevel-doc-formality} and \ref{fig:sentLevel-conv-formality} illustrate the formality change results for document-level and conversation-level datasets, respectively. Figures \ref{fig:sentLevel-doc-sentiment} and \ref{fig:sentLevel-conv-sentiment} display the sentiment transfer results for document-level and conversation-level datasets, respectively. Figures \ref{fig:ctxtLevel-ccc-toxicity}, \ref{fig:sentLevel-mdmd-toxicity}, and \ref{fig:sentLevel-prosocial-toxicity} depict the de-toxification results for conversational datasets. Notably, all of these figures exhibit similar trends to the aggregate results across all tasks and datasets presented in Figure \ref{fig:sent_auto_agg}. 

\subsection{Correlation with Human Judgments} 
\label{app:ssec:sent-auto-corr}
Effective evaluation metrics should yield judgments that correlate highly with human judgments, assuming
that human evaluators represent a gold-standard. For the human judgments along the dimensions of naturalness and fit, we map human preferences as follows: `contextual' to $1$, `tie' to 0, and `non-contextual' to $-1$. For the automatic metrics, we assign a score of $1$ if a metric scores the contextual rewrite higher than the non-contextual rewrite, and $-1$ if the metric scores are lower for contextual rewrites. 

For a given automatic metric and human judgment dimension, we calculate the Spearman rank $\rho$ correlation and Kendall's $\tau$ for the dataset samples used during the contextual human evaluation \S\ref{ssec:human_eval_setup}. The correlation scores, ranging from $-1$ to $1$, are obtained by comparing the mapped automatic scores with the mapped human judgment scores. Higher values indicate a stronger correlation between the scores obtained using the comparison metric and judgments made by human evaluators.
Refer to Tables \ref{tab:toxicity_corr_sent} -- \ref{tab:sentiment_corr_sent} for the correlation scores of non-contextual evaluation metrics with human judgments for each task.
\begin{table}[H]
    \centering
   \resizebox{1.\columnwidth}{!}{\begin{tabular}{ccccc}
          & Lexical ($\rho$) & Semantic ($\rho$) & Lexical ($\tau$) & Semantic ($\tau$) \\
        \toprule
        Fit &  -0.02&0.14&-0.02&0.14 \\
        \midrule
        Naturalness & -0.03	& 0.18&-0.03&0.17 \\
    \end{tabular}}
    \caption{Detoxification task: Spearman rank and Kendall Correlation of non-contextual evaluation metrics with human judgment}
    \label{tab:toxicity_corr_sent}
\end{table}
\begin{table}[H]
    \centering
   \resizebox{1.\columnwidth}{!}{\begin{tabular}{ccccc}
          & Lexical ($\rho$) & Semantic ($\rho$) & Lexical ($\tau$) & Semantic ($\tau$) \\
        \toprule
        Fit &  0.18	& 0.28 & 0.17& 0.27 \\
        \midrule
        Naturalness & 0.11 & 0.26 & 0.10 & 0.24 \\
    \end{tabular}}
    \caption{Formality task: Spearman rank and Kendall Correlation of non-contextual evaluation metrics with human judgment}
    \label{tab:formality_corr_sent}
\end{table}

\begin{table}[H]
    \centering
   \resizebox{1.\columnwidth}{!}{\begin{tabular}{ccccc}
          & Lexical ($\rho$) & Semantic ($\rho$) & Lexical ($\tau$) & Semantic ($\tau$) \\
        \toprule
        Fit &  0.11	& 0.26&0.10&0.25 \\
        \midrule
        Naturalness & -0.05&0.13&-0.05&0.12 \\
    \end{tabular}}
    \caption{Sentiment task: Spearman rank and Kendall Correlation of non-contextual evaluation metrics with human judgment}
    \label{tab:sentiment_corr_sent}
\end{table}

\begin{table}[H]
    \centering
   \resizebox{1.\columnwidth}{!}{\begin{tabular}{ccccccc}
          & Lexical ($\rho$) & Semantic ($\rho$) & \ourCustomMetric ($\rho$) & Lexical ($\tau$) & Semantic ($\tau$) & \ourCustomMetric ($\tau$) \\
        \toprule
        Fit &  0.63&0.56&0.85&0.61&0.54&0.82 \\
        \midrule
        Naturalness & 0.59&0.58&0.88&0.56&0.55&0.84 \\
    \end{tabular}}
    \caption{Detoxification task: Spearman rank and Kendall Correlation of contextual evaluation metrics with human judgment}
    \label{tab:toxicity_corr_ctxt}
\end{table}

\begin{table}[H]
    \centering
   \resizebox{1.\columnwidth}{!}{\begin{tabular}{ccccccc}
          & Lexical ($\rho$) & Semantic ($\rho$) & \ourCustomMetric ($\rho$) & Lexical ($\tau$) & Semantic ($\tau$) & \ourCustomMetric ($\tau$) \\
        \toprule
        Fit &  0.74&0.68&0.93&0.71&0.65&0.89 \\
        \midrule
        Naturalness & 0.68&0.69&0.94&0.65&0.66&0.90 \\
    \end{tabular}}
    \caption{Formality task: Spearman rank and Kendall Correlation of contextual evaluation metrics with human judgment}
    \label{tab:formality_corr_ctxt}
\end{table}

\begin{table}[H]
    \centering
   \resizebox{1.\columnwidth}{!}{\begin{tabular}{ccccccc}
          & Lexical ($\rho$) & Semantic ($\rho$) & \ourCustomMetric ($\rho$) & Lexical ($\tau$) & Semantic ($\tau$) & \ourCustomMetric ($\tau$) \\
        \toprule
        Fit &  0.45&0.45&0.78&0.42&0.52&0.74 \\
        \midrule
        Naturalness & 0.44&0.51&0.73&0.42&0.48&0.69 \\
    \end{tabular}}
    \caption{Sentiment task: Spearman rank and Kendall Correlation of contextual evaluation metrics with human judgment}
    \label{tab:sentiment_corr_ctxt}
\end{table}

\begin{table*}[b]
\centering
\includegraphics[width=2\columnwidth]{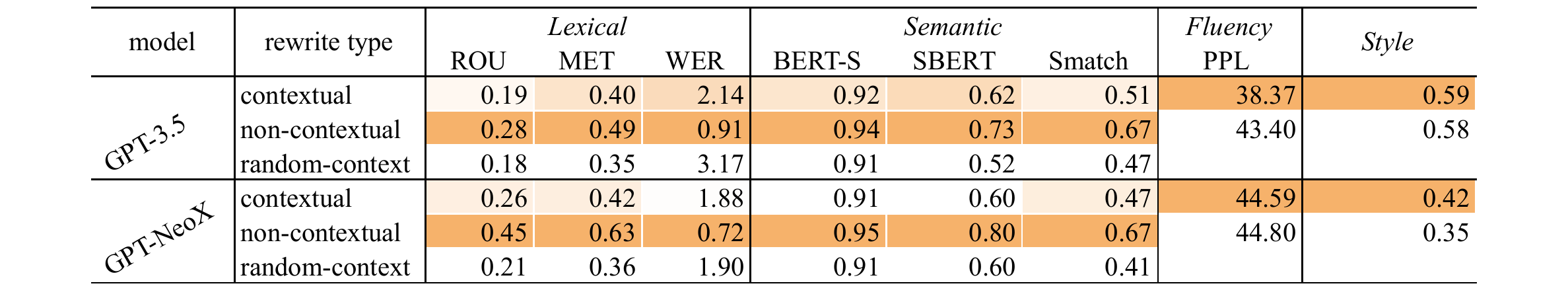} 
\caption{Non-contextual Automatic Evaluation Results on \textbf{Formality}: Document-level context from CNN/DailyMail + Blog Authorship Corpus }
\label{fig:sentLevel-doc-formality}
\end{table*}

\begin{table*}[b]
\centering
\includegraphics[width=2\columnwidth]{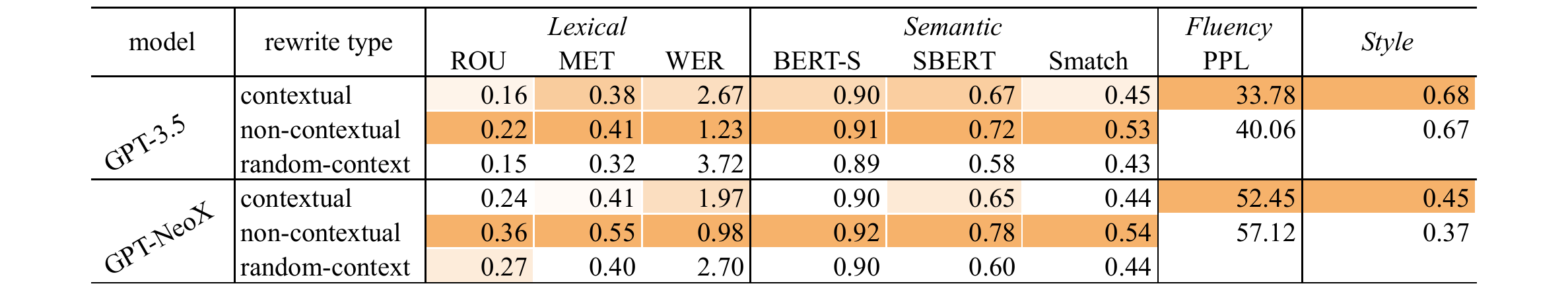} 
\caption{Non-contextual Automatic Evaluation Results on \textbf{Formality}: Conversational context comprised of Reddit threads }
\label{fig:sentLevel-conv-formality}
\end{table*} 

\begin{table*}[t]
\centering
\includegraphics[width=2\columnwidth]{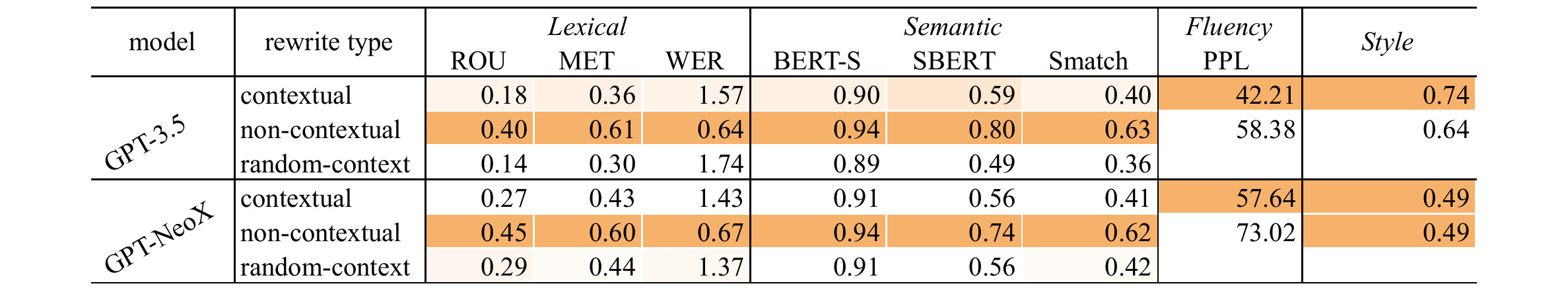} 
\caption{Non-contextual Automatic Evaluation Results on \textbf{Sentiment}: Document-level context comprised of Yelp Reviews }
\label{fig:sentLevel-doc-sentiment}
\end{table*}

\begin{table*}[t]
\centering
\includegraphics[width=2\columnwidth]{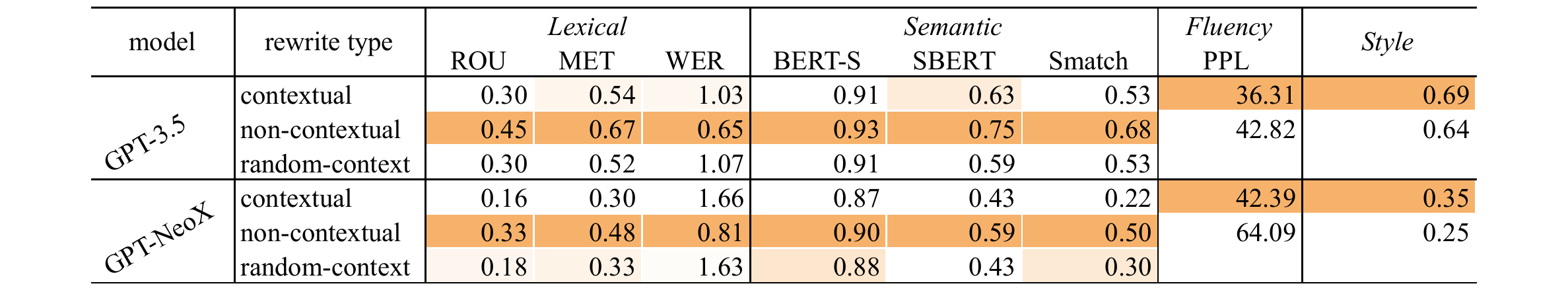} 
\caption{Non-contextual Automatic Evaluation Results on \textbf{Sentiment}: Conversational context from DailyDialog dataset }
\label{fig:sentLevel-conv-sentiment}
\end{table*} 

\begin{table*}[t]
\centering
\includegraphics[width=2\columnwidth]{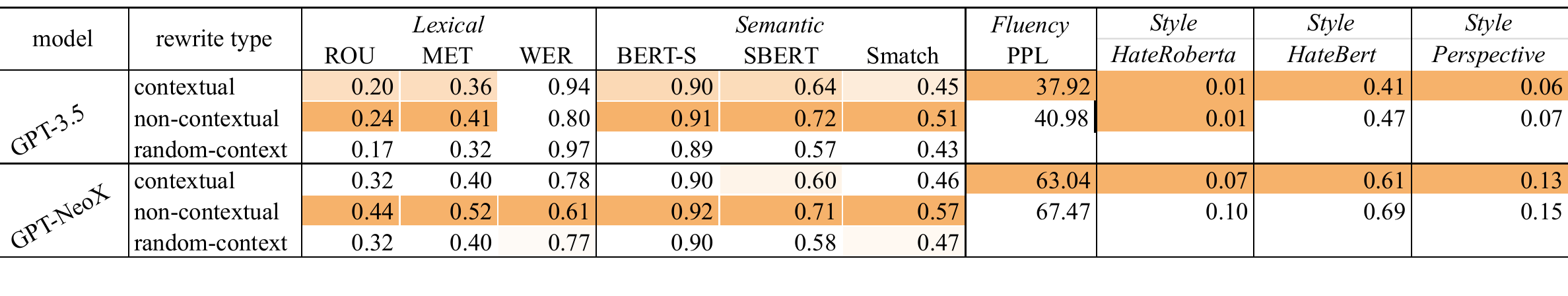} 
\caption{Non-contextual Automatic Evaluation Results on \textbf{Toxicity}: Conversational context from CCC dataset}
\label{fig:sentLevel-ccc-toxicity}
\end{table*} 

\begin{table*}[t]
\centering
\includegraphics[width=2\columnwidth]{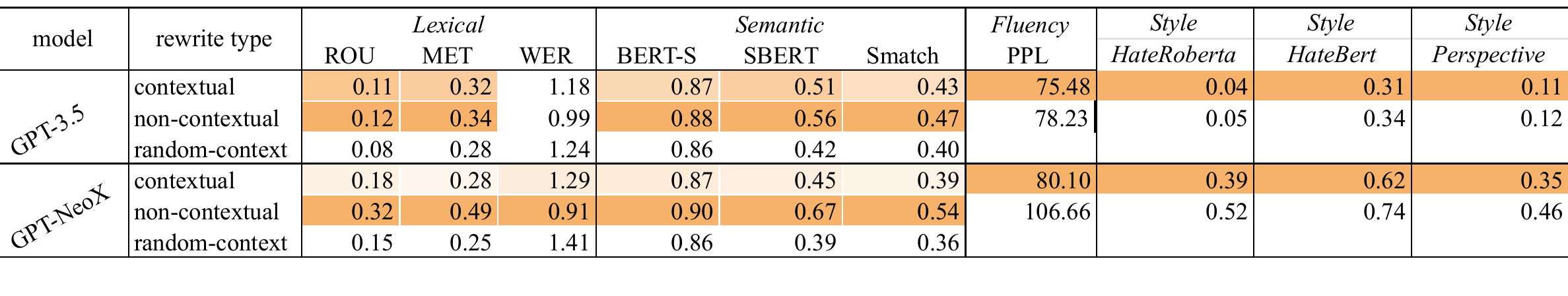} 
\caption{Non-contextual Automatic Evaluation Results on \textbf{Toxicity}: Conversational context from MDMD dataset}
\label{fig:sentLevel-mdmd-toxicity}
\end{table*} 

\begin{table*}[t]
\centering
\includegraphics[width=2\columnwidth]{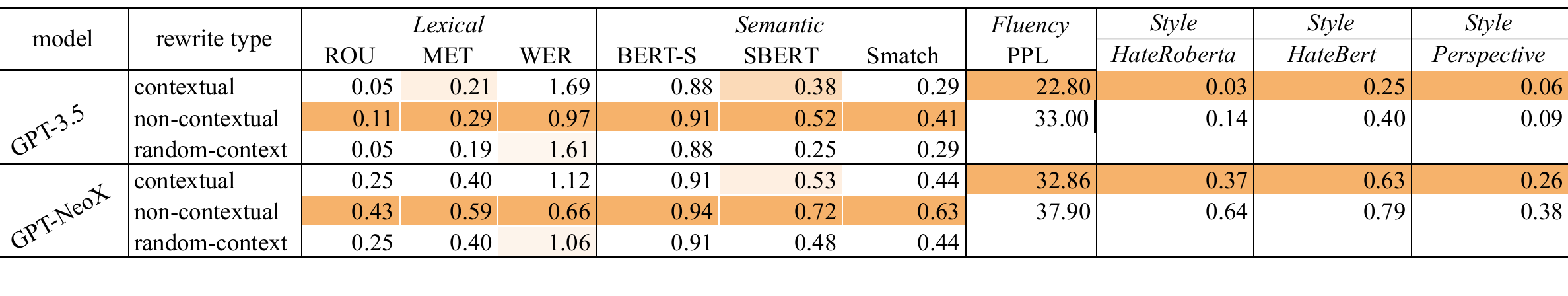} 
\caption{Non-contextual Automatic Evaluation Results on \textbf{Toxicity}: Conversational context from ProsocialDialog dataset}
\label{fig:sentLevel-prosocial-toxicity}
\end{table*}

\section{Contextual Automatic Evaluation}
\label{sec:app:ctxt-auto}
We present the contextual automated evaluation results for each task-specific dataset. Figures \ref{fig:ctxtLevel-doc-formality} and \ref{fig:ctxtLevel-conv-formality} illustrate the formality change results for document-level and conversation-level datasets, respectively. Figures \ref{fig:ctxtLevel-doc-sentiment} and \ref{fig:ctxtLevel-conv-sentiment} display the sentiment transfer results for document-level and conversation-level datasets, respectively. Figures \ref{fig:ctxtLevel-ccc-toxicity}, \ref{fig:ctxtLevel-mdmd-toxicity}, and \ref{fig:ctxtLevel-prosocial-toxicity} depict the de-toxification results for conversational datasets. All of these figures exhibit similar trends to the aggregate results across all tasks and datasets presented in Figure \ref{fig:sent_auto_agg} and they align with the findings from our contextual human evaluation study.  

\subsection{Correlation with Human Judgments}
\label{app:ssec:ctxt-auto-corr}
Similar to \S\ref{app:ssec:sent-auto-corr}, we measure the Spearman rank $\rho$ correlation and Kendall's $\tau$ correlation for the samples used during human evaluation in \S\ref{ssec:human_eval_setup}. 
Refer to Tables \ref{tab:toxicity_corr_ctxt} -- \ref{tab:sentiment_corr_ctxt} for the correlation scores of non-contextual evaluation metrics with human judgments for each task.

\begin{table*}[!htb]
\centering
\includegraphics[width=2\columnwidth]{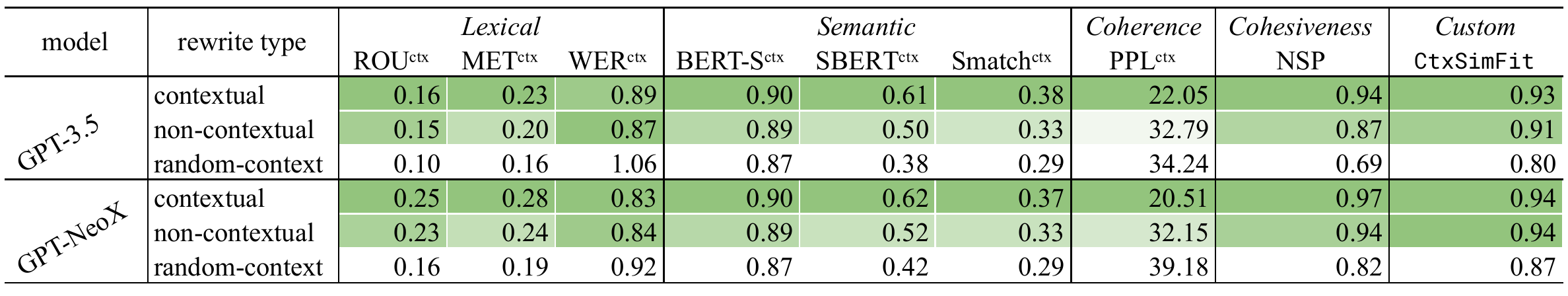} 
\caption{Contextual Automatic Evaluation Results on \textbf{Formality}: Document-level context from CNN/DailyMail + Blog Authorship Corpus }
\label{fig:ctxtLevel-doc-formality}
\end{table*}

\begin{table*}[!htb]
\centering
\includegraphics[width=2\columnwidth]{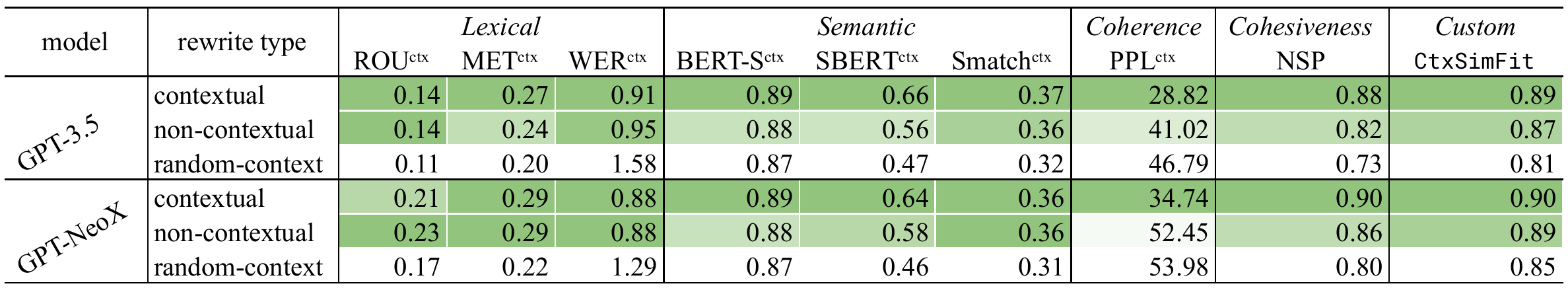} 
\caption{Contextual Automatic Evaluation Results on \textbf{Formality}: Conversational context comprised of Reddit threads }
\label{fig:ctxtLevel-conv-formality}
\end{table*} 

\begin{table*}[!htb]
\centering
\includegraphics[width=2\columnwidth]{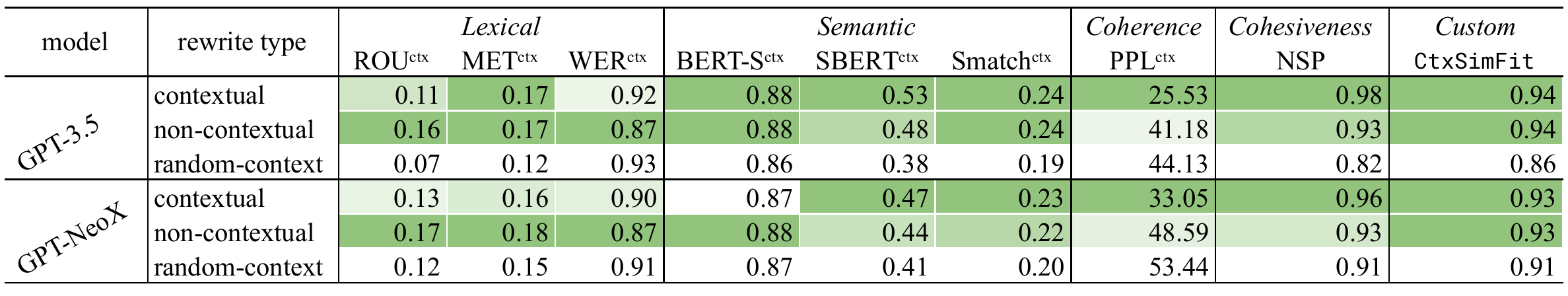} 
\caption{Contextual Automatic Evaluation Results on \textbf{Sentiment}: Document-level context comprised of Yelp Reviews }
\label{fig:ctxtLevel-doc-sentiment}
\end{table*}

\begin{table*}[!htb]
\centering
\includegraphics[width=2\columnwidth]{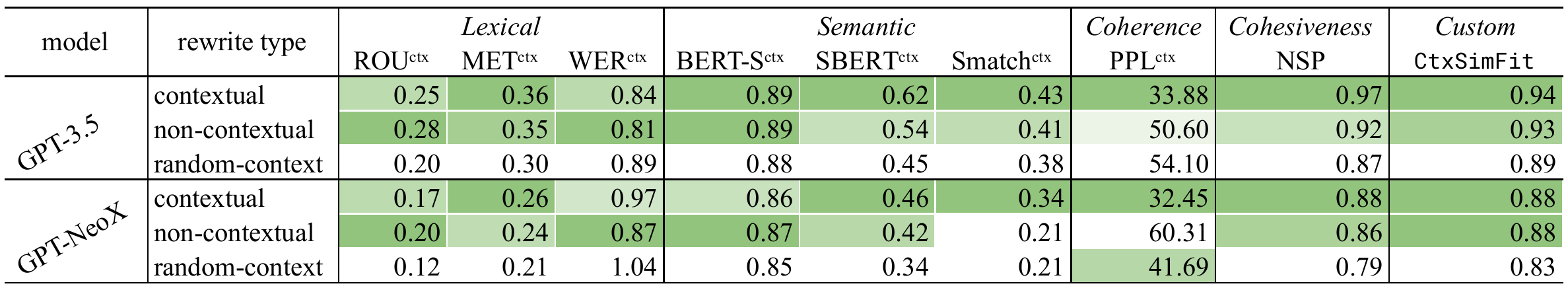} 
\caption{Contextual Automatic Evaluation Results on \textbf{Sentiment}: Conversational context from DailyDialog dataset }
\label{fig:ctxtLevel-conv-sentiment}
\end{table*} 

\begin{table*}[!htb]
\centering
\includegraphics[width=2\columnwidth]{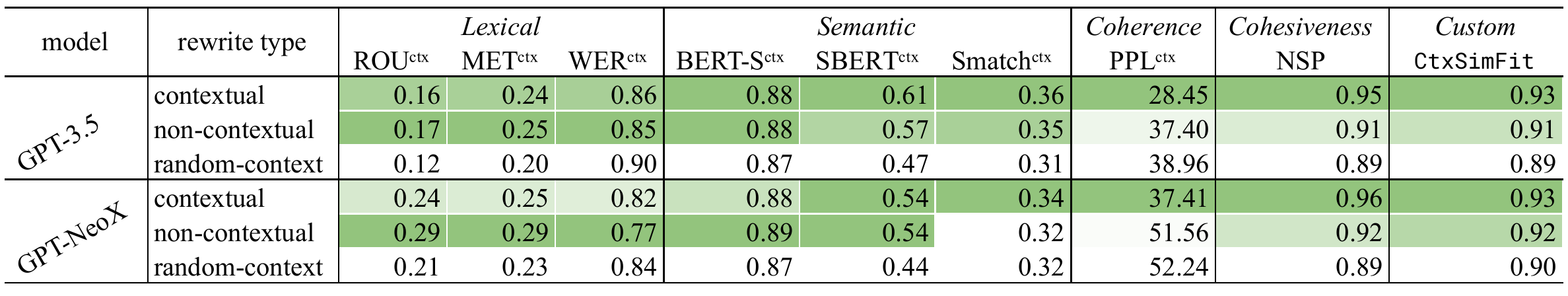} 
\caption{Contextual Automatic Evaluation Results on \textbf{Toxicity}: Conversational context from CCC dataset}
\label{fig:ctxtLevel-ccc-toxicity}
\end{table*} 

\begin{table*}[!htb]
\centering
\includegraphics[width=2\columnwidth]{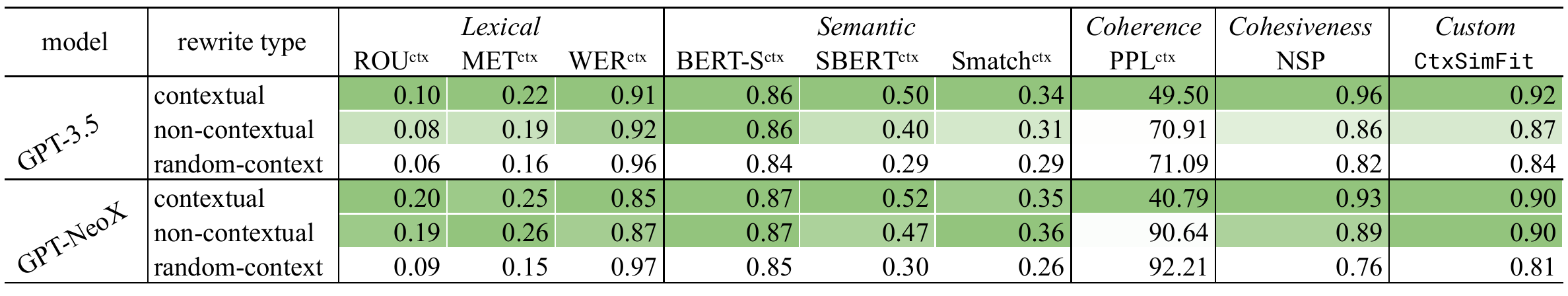} 
\caption{Contextual Automatic Evaluation Results on \textbf{Toxicity}: Conversational context from MDMD dataset}
\label{fig:ctxtLevel-mdmd-toxicity}
\end{table*} 

\begin{table*}[!htb]
\centering
\includegraphics[width=2\columnwidth]{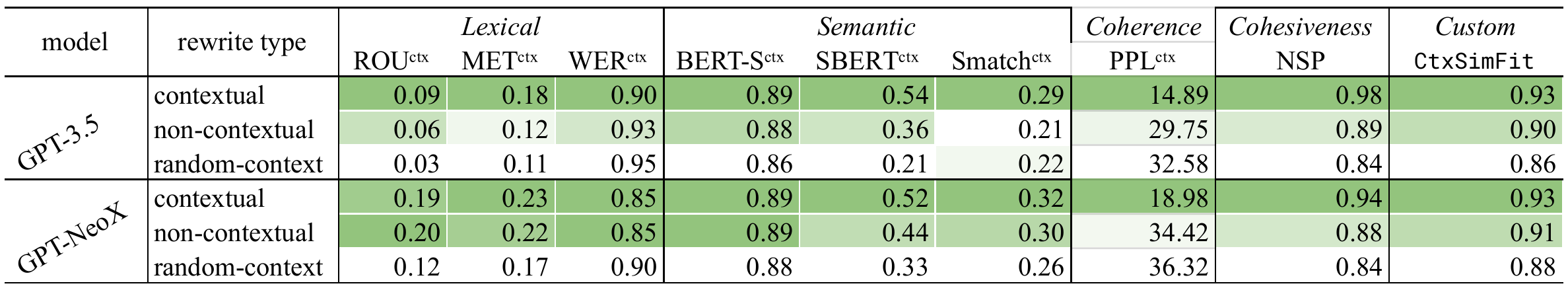} 
\caption{Contextual Automatic Evaluation Results on \textbf{Toxicity}: Conversational context from ProsocialDialog dataset}
\label{fig:ctxtLevel-prosocial-toxicity}
\end{table*}

\end{document}